\documentclass[11pt]{article}

\usepackage[final]{acl}

\usepackage{times}
\usepackage{latexsym}

\usepackage[T1]{fontenc}

\usepackage[utf8]{inputenc}

\usepackage{microtype}

\usepackage{inconsolata}

\usepackage{graphicx}

\usepackage[shortlabels,inline]{enumitem}
\usepackage{tikz,lipsum}
\usepackage[most]{tcolorbox}
\usepackage{verbatim}
\usepackage{amssymb}
\usepackage{minted}
\usepackage{fancyvrb}
\usepackage{ragged2e}
\usepackage[dvipsnames]{xcolor}
\usepackage{amsmath}
\usepackage{booktabs}
\usepackage{tabularray}
\usepackage{arydshln}
\usepackage{stmaryrd}
\usepackage{marvosym}
\usepackage{makecell}
\usepackage{multirow}
\usepackage{longtable}
\usepackage{adjustbox}
\usepackage{dsfont}
\usepackage{xspace}

\usepackage{cleveref}
\crefname{section}{\S}{\S}
\crefname{table}{Table}{Tables}
\crefname{figure}{Fig.}{Figs.}
\crefname{algorithm}{Alg.}{}
\crefname{ALC@unique}{Line}{Lines}
\crefname{equation}{Eq.}{Eqs.}
\crefname{appendix}{App.}{Apps.}
\crefformat{appendix}{App.~#2#1#3}
\crefformat{section}{\S#2#1#3}
\usepackage{soul}


\usepackage{todonotes}

\NewDocumentCommand{\prompt}{O{} +m}{%
\begin{center}
\begin{tcolorbox}[
    width=0.94\columnwidth,
    coltitle=white,
    colframe=black,
    colback=black!5!white,
    boxrule=1pt,
    enhanced jigsaw,
    breakable,
    pad at break*=2mm,
    left=2pt,
    right=2pt,
    top=2pt,
    bottom=2pt,
    fontupper=\small,
    fontlower=\small,
    title={#1},
]
#2
\end{tcolorbox}
\end{center}
}

\newcommand{\taubench}{$\tau$-Bench\xspace}

\RequirePackage{etoolbox}
\RequirePackage{pgf}
\RequirePackage[table]{xcolor}

\definecolor{high}{HTML}{B5CCF7}
\definecolor{low}{HTML}{FFFFFF}

\newcommand{\gca}[1]{#1}

\newtcolorbox[auto counter,number within=section]{promptbox}[2][]{
    coltitle=white,
    colframe=black,
    colback=black!5!white,
    enhanced jigsaw,
    breakable,
    pad at break*=2mm,
    fontupper=\footnotesize,
    fontlower=\footnotesize,
    fonttitle=\small,
    left=4pt,
    right=4pt,
    title={Prompt \thetcbcounter: #2},
    label={#1},
    label type=promptbox,
}
\crefname{promptbox}{Prompt}{Prompts}
\Crefname{promptbox}{Prompt}{Prompts}

\newcommand{\oursone}{\textbf{\texttt{PA\textsuperscript{1}}}\xspace}

\newcommand{\ourstwo}{\textbf{\texttt{PA\textsuperscript{2}}}\xspace}

\newcommand{\ours}{\textbf{\texttt{PA\textsuperscript{3}}}\xspace}

\title{\ours: \textbf{P}olicy-\textbf{A}ware \textbf{A}gent \textbf{A}lignment through Chain-of-Thought}

\author{
  \textbf{Shubhashis Roy Dipta}\thanks{Work done during an internship at Amazon Alexa AI}\textsuperscript{1},
  \textbf{Daniel Bi\'{s}}\textsuperscript{2},
  \textbf{Kun Zhou}\textsuperscript{2},
  \textbf{Lichao Wang}\textsuperscript{2},\\[0.3em]
  \textbf{Benjamin Yao}\textsuperscript{2},
  \textbf{Chenlei Guo}\textsuperscript{2},
  \textbf{Ruhi Sarikaya}\textsuperscript{2}
  \\[0.8em]
  \textsuperscript{1}University of Maryland, Baltimore County \quad
  \textsuperscript{2}Amazon Alexa AI
  \\[0.4em]
  \small{\textbf{Correspondence:} \href{mailto:sroydip1@umbc.edu}{sroydip1@umbc.edu}}
}

\begin{document}
\maketitle

\begin{abstract}

Conversational assistants powered by large language models (LLMs) excel at tool-use tasks but struggle with adhering to complex, business-specific rules. While models can reason over business rules provided in context, including all policies for every query introduces high latency and wastes compute. Furthermore, these lengthy prompts lead to long contexts, harming overall performance due to the `needle-in-the-haystack' problem.
To address these challenges, 
we propose a multi-stage alignment method that teaches models to recall and apply relevant business policies during chain-of-thought reasoning at inference time, \textit{without including the full business policy in-context}.
Furthermore, we introduce a novel \textbf{\textit{P}olicy\textit{R}ecall} reward based on the Jaccard score and a Hallucination Penalty for GRPO training. Altogether, our best model outperforms the baseline by \textbf{16 points} and surpasses comparable \textit{in-context} baselines of similar model size by \textbf{3 points}, while using \textbf{40\% fewer words}.

\end{abstract}

\section{Introduction}

\begin{figure}
    \centering
    \includegraphics[width=1\linewidth]{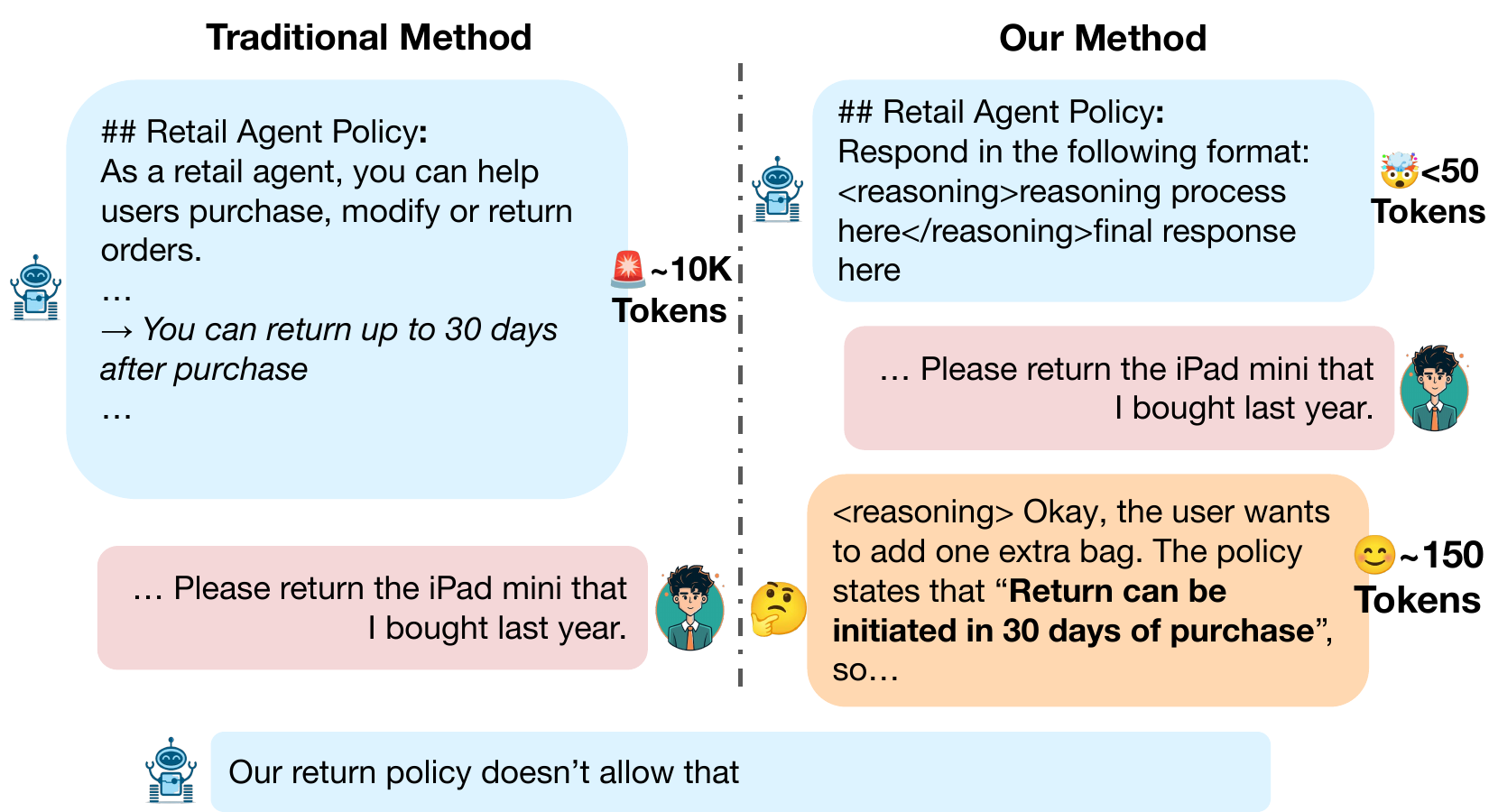}
    \caption{While traditional policy-adherence agents need in-context business policies which can range from \textbf{10k to 90k tokens}, our method only recalls the relevant policies. For each request, the relevant policies can extend up to \textbf{150-400 tokens}, which uses up to \textbf{225x fewer} policy tokens per request than the traditional method.}
    \label{fig:intro}
    \vspace{-4mm}
\end{figure}

Different businesses implement different business policies. For example, one e-commerce platform can provide a 30-day return window, while the other might offer only 15 days. Such variation makes developing a single agent model infeasible. While traditional methods, as described on the left of \cref{fig:intro}, focus on providing in-context business policies, the business policies can range from 10k to 90k tokens, which can significantly increase costs, latency and throughput. \taubench \citep{yao2024tau}, a policy-based tool-calling benchmark, has shown that 95.9\% of inference costs come from input tokens, with only 4.1\% from output. In this work, we explore whether we can eliminate all business policies from in-context by teaching models to recall business policies during inference. As illustrated on the right of \cref{fig:intro}, we show that our aligned model \textbf{recalls ONLY the relevant policies} during its reasoning, and adheres to them. For each assistant turn, the number of relevant policies can be as low as 0 (e.g., a final ``Thank You''), and in most cases at most 1-5 policies, for example, the agent only needs the \textit{return-policy} to initiate a return. By removing the full business policy from in-context, our method reduces the total number of words by 40\%. Finally, \textbf{we propose a structured alignment recipe, adaptable to new business domains with explicit policy documents}.

Specifically, our model learns the business policies through reasoning steps and later during inference, recall only the relevant business policies. For any given conversation history and business policy document, first, we generate chain-of-thought targeted to relevant business policies through an evaluation-filtering cycle. Next, we use those CoT traces with multistage training and teach the model to recall the relevant policies during inference and adhere to those. \citet{guan_deliberative_2025} showed that safety knowledge can be injected through CoT traces. However, their method assumes that they have access to high-quality CoTs, whereas our method generates those high-quality CoT through a novel evaluation-filtering cycle, hence it can be extended to any business use case. We have shown that this evaluation-filtering cycle during CoT generation improves the model's recall capability. Additionally, our method is trained with a novel policy-recall reward that reduces hallucinations while encouraging shorter and more focused policy recalls.
To sum up, 
\begin{itemize}
    \item We propose a generate-branch-evaluate-refine method that extracts and filters \textbf{Complete and Reliable CoTs} automatically, so the reasoning traces do not have to be written by hand. It adapts to other domains with explicit written policies.
    \item We propose a novel policy-recall-based reward and hallucination-based penalty for training policy-adherent agents.
    \item Our best model shows \textbf{~16 points improvement} over the no-business-policy baseline and \textbf{~3 points improvement} over the in-context business-policy baseline while using \textbf{40\% fewer words}.
\end{itemize}

\section{Related Work}
\subsection{Function Calling Dataset}
Many datasets have been proposed for training and evaluating function-calling capabilities. Recently, \citet{yao2024tau} developed \taubench evaluation benchmark based on multi-domain business policy. Later, \citet{prabhakar_apigen-mt_2025} extended the \taubench to a train dataset, using the idea of \citet{liu2024apigen}. \citet{berkeley-function-calling-leaderboard} have also developed a multi-turn, multi-step evaluation dataset, but it lacks the business policy that is relevant to real-life use cases. Recently, \citet{acikgoz_can_2025} have published a mix of function calling and intent detection dataset, and \citet{xu2026patternawaretoolintegratedreasoning} study \textit{how}, not just \textit{when}, tools are applied.

\subsection{Prompt Compression}
Prior work has extensively explored compressing Chain-of-Thought (CoT) to reduce latency and generation costs \citep{gu-etal-2025-text, su_token_2025, li-etal-2023-compressing}. \citet{cheng_compressed_2024} learn compressed, continuous latent embeddings that are much shorter than the original CoT tokens, while \citet{su_token_2025} use VQ-VAE to map reasoning paths to discrete latent tokens and then train models on downstream tasks using those tokens. In contrast, our method focuses on compressing the system prompt rather than the reasoning path. During inference, it recalls only a small set of relevant policies from long policy documents, thereby shortening the policy context.

\subsection{Chain-of-Thought Evaluation}
Although there has been relatively little work on CoT evaluation itself, there is extensive research on evaluating generated text. \citet{fu-etal-2024-gptscore,roydipta2026vcinspector} have shown how LLMs can be used as a judge for the generative text. Later, \citet{chiang_closer_2023} have shown that this idea can be extended and can be improved using an analysis-based prompting rather than direct scoring. \citet{wang-etal-2024-large-language-models-fair} have shown that LLMs are neither consistent nor fair evaluators but provided recipes to overcome that inconsistency, as do \citet{roydipta2025depresuppose} for prompt variance. Recently, \citet{saha-etal-2024-branch} and \citet{li-etal-2025-dna} have provided a decomposition and aggregation based evaluation which outperformed the previous works. \citet{wang2022self} and \citet{lee-etal-2024-tree} have shown that generating multiple reasoning paths from the same model and aggregating them can substantially improve performance.
Our method draws inspiration from these works but focuses on evaluating intermediate reasoning steps using various rubrics, which outcome-only evaluation misses \citep{mazumder2026agentcollabbench}.

\subsection{Verifiable Rewards in GRPO}
To address the high computational cost associated with early reinforcement learning algorithms \citep{ouyang2022traininglanguagemodelsfollow} such as PPO \citep{schulman2017proximalpolicyoptimizationalgorithms}, \citet{shao_deepseekmath_2024} introduced Group-Relative Policy Optimization (GRPO), a lightweight RL framework that has achieved strong alignment performance across multiple domains, including mathematical reasoning \citep{shao_deepseekmath_2024, wu_confucius3-math_2025} and general reasoning tasks \citep{deepseek-ai_deepseek-r1_2025, zheng_group_2025}. GRPO is paired with correctness-based rewards or LLM-as-a-judge scoring \citep{gu2025surveyllmasajudge}.
More recently, GRPO has been adapted to a broader set of domains through domain-specific, verifiable reward formulations. For example, \citet{qian_toolrl_2025} employ function-matching scores as rewards for training tool-calling LLMs, \citet{he_veri-r1_2025} combine final verification accuracy with retrieval metrics to improve claim verification, \citet{dipta2026decomposerl} use an ensemble of complementary scores, and \citet{tennant2025moralalignmentllmagents} incorporate human moral values directly into the reward function to align ethical behavior.
Our work extends this line of verifiable reward design by introducing a novel policy-recall metric that rewards correct policy recall while penalizing incorrect or hallucinated policy recalls, enabling more faithful and policy-grounded reasoning.

\section{Method}
We propose a recipe with two parts: (1) Generating \textit{High-Quality} Chain-of-Thought (\cref{sec:cot_generation}) (2) Multi-Stage training using the CoT data (\cref{sec:multi_stage_training}).

\subsection{Chain-of-Thought} \label{sec:cot_generation}
Unlike the traditional Chain-of-Thought (CoT) that thinks step by step, we want our CoTs to recall only the relevant policies that are related to the user request and adhere to those policies during the final response. With that in mind, we use  a \textbf{generate-branch-evaluate-refine} -- a 4 stage cycle to create our policy-based CoTs. The whole pipeline is shown in \cref{fig:cot_generation}.

\subsubsection{Stage 1: Generate}
For each assistant or tool turn, we prompt an LLM to generate the CoT. Specifically, given the business policy, and the whole user-assistant conversation up to that turn, we prompt the model to generate the CoT only for the last user-assistant turn. We provide the full conversation as in some cases it might be necessary to get the context, i.e., the user has already provided the user ID and the agent does not need the policy to ``ask for user ID'' again. Unless specified otherwise, we use \texttt{deepseek-r1} to generate the CoTs.

\subsubsection{Stage 2: Rubrics}
While one trivial approach would be to just trust the LLM and take the CoT as generated, there is a high risk of noise, i.e., made-up policy, non-relevant policy, or missing out important-relevant policy. Inspired by the previous works \citep{saha-etal-2024-branch, lanham2023measuringfaithfulnesschainofthoughtreasoning}, we branch the evaluation space into 4 different evaluator agents. We identify four core rubrics of the expected CoTs: Atomicity, Completeness, Faithfulness, Style.

\begin{figure*}[!t]
    \centering
    \begin{minipage}[b]{0.48\textwidth}
        \centering
        \includegraphics[width=0.99\linewidth]{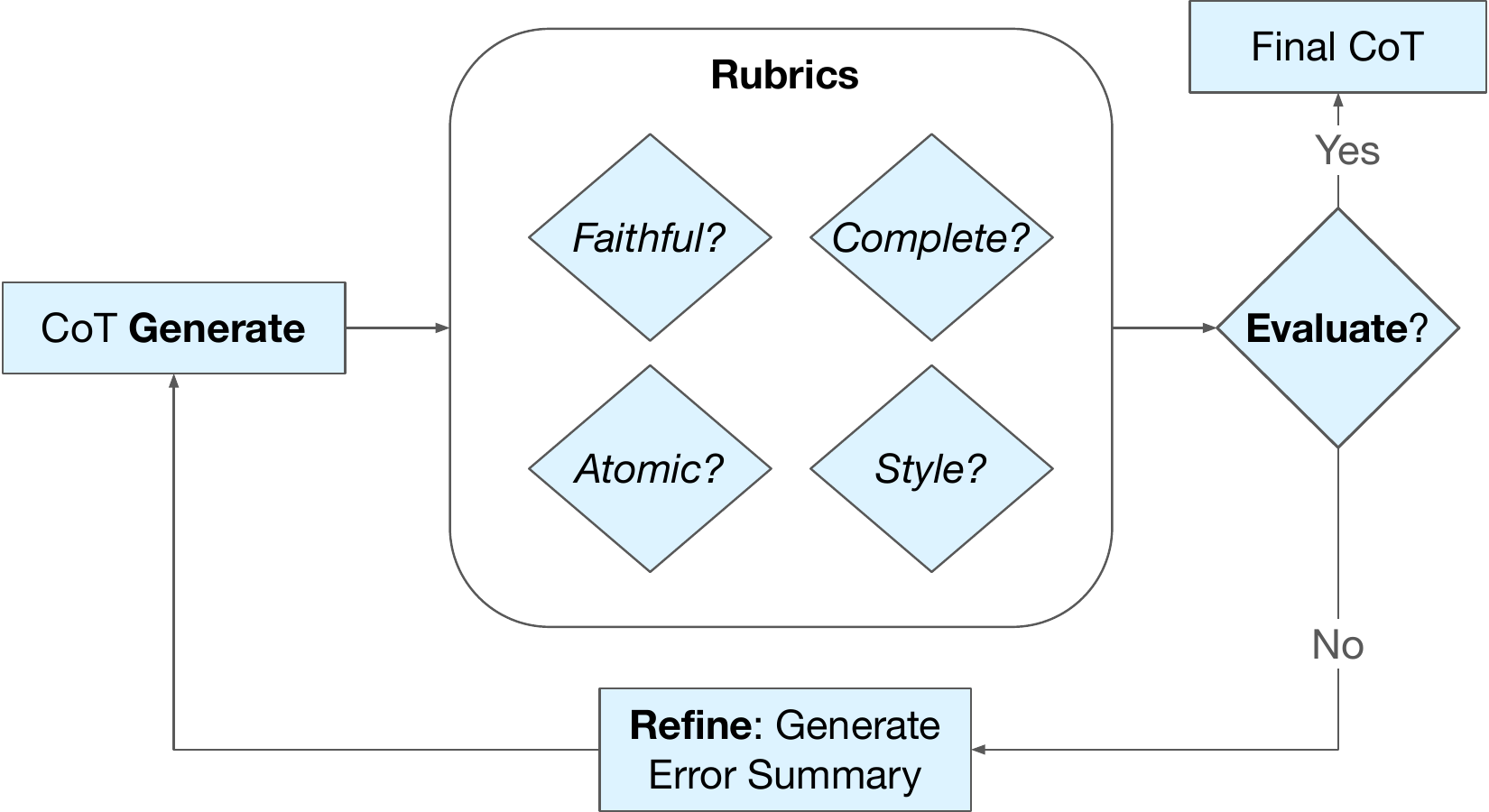}
        \caption{Overview of our multi-stage CoT refinement loop, consisting of Generation, Rubric Evaluation, CoT Evaluation, Targeted Refinement.}
        \label{fig:cot_generation}
    \end{minipage}\hfill
    \begin{minipage}[b]{0.48\textwidth}
        \centering
        \includegraphics[width=0.99\linewidth]{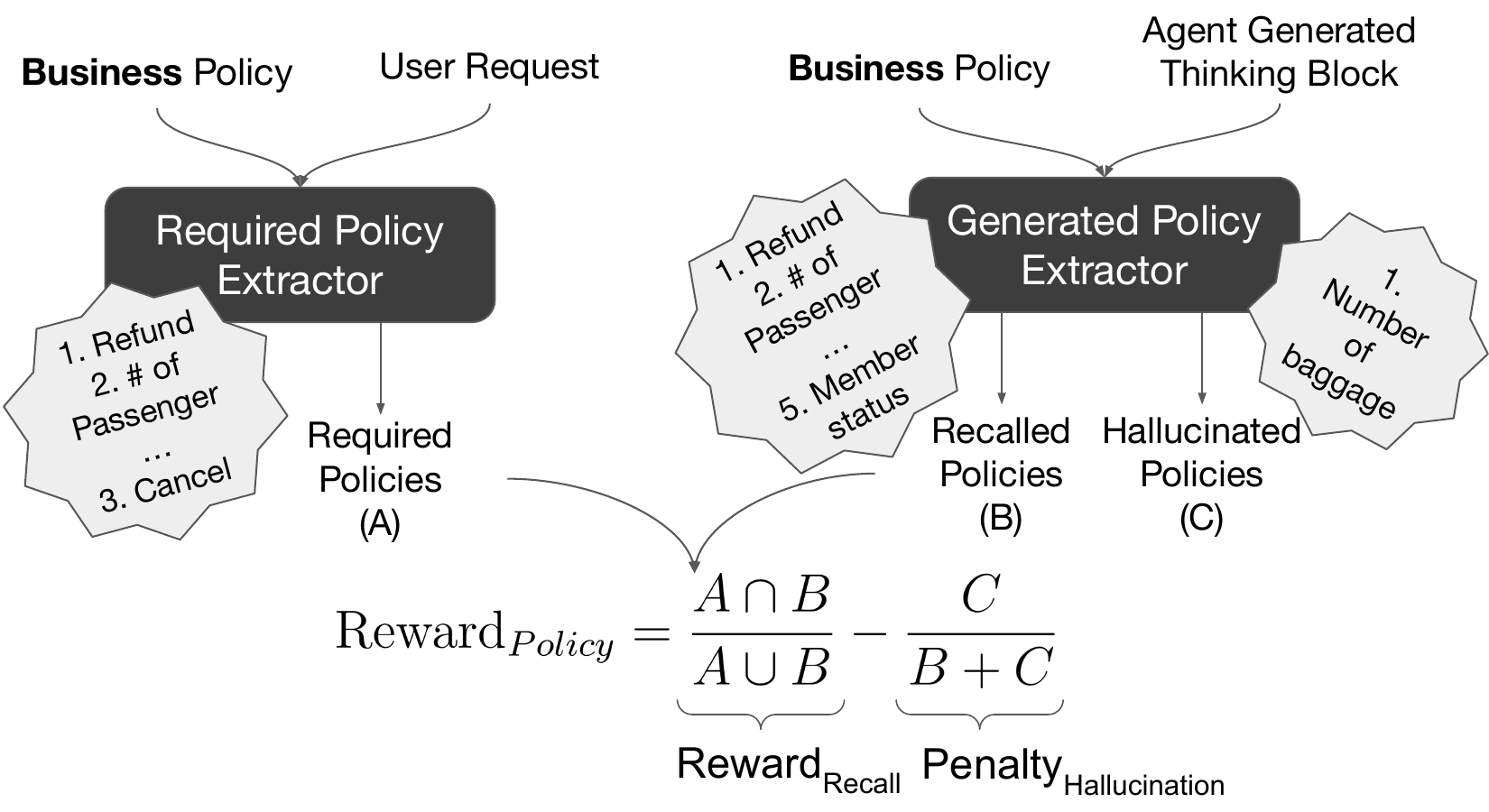}
        \caption{Overview of the proposed \textbf{\textit{P}olicy\textit{R}ecall} reward, consisting of a policy-recall–based reward and a hallucination-based penalty.}
        \label{fig:policy_recall_reward}
    \end{minipage}
    \vspace{-3mm}
\end{figure*}

\textbf{Completeness:} The CoT must include all policies that are relevant to satisfying the user request.

\textbf{Atomicity:} The CoT must be concise such that it does not include irrelevant policies.

\textbf{Faithfulness:} The CoT must not mention anything that is not explicitly stated on the policy.

\textbf{Style:} The CoT should have a thinking narrative style rather than just extraction from the document. 

\subsubsection{Stage 3: Evaluate}
We prompt each of the evaluators to score depending on their own properties as described above. Following \citet{chiang_closer_2023}, we use an ``analyze-rate'' prompt, and we score on the 1-10 scale used by \citet{stureborg_large_2024}. We use \texttt{Claude-3.5-v1} as the evaluator unless otherwise specified. We hand-engineer the thresholds based on the principle that 
\textit{Faithfulness is the most important property}, followed in order 
by Completeness, Atomicity, and Style. Accordingly, Faithfulness 
requires a perfect score of 10, while the remaining rubrics use 
progressively lower thresholds (details in App.~\ref{app:thresholds}). 
A CoT must satisfy all thresholds simultaneously; otherwise, it is 
sent to the refinement stage.

\subsubsection{Stage 4: Refinement}
The CoTs that have not passed the thresholds are sent to a summarizer agent which summarizes what the CoT Generator has done right and what it has done wrong. This concise summary goes to the CoT Generator again to refine the previous CoT based on the evaluation summary. We attempt at most four refinement rounds, as we have seen diminishing returns after that (\cref{fig:cot_analysis}). If not found by then, we remove that data point entirely.
Prompts for the CoT generation are provided in \cref{app:prompts_for_cot_data}.

\subsection{Multi-Stage Training} \label{sec:multi_stage_training}
We use a three-stage training to inject the business knowledge into the parametric knowledge of the LLM. We use the \texttt{Qwen2.5-Instruct-32B} as our base model. A high level overview of the multi-stage training framework is shown in \cref{tab:training-stages} in the appendix.

\subsubsection{Stage 1: Continual-SFT} \label{sec:stage_1}
In the first stage, we use a mix of general function calling dataset and business-policy in-context dataset to fine-tune the base model. The main goal of this stage was to improve the general function calling capability and train on the business-policy-adherent tool calling. For the business-policy based function-calling dataset, we keep the business policy in-context for this stage. We refer to the resulting model as \oursone.

\subsubsection{Stage 2: CoT Augmented SFT} \label{sec:stage_2}
In the next stage, we use the final checkpoint from the stage 1 as the base model and we continue fine-tuning using the CoT-augmented business-policy dataset. This dataset was synthetically generated using the CoT extraction pipeline (\cref{sec:cot_generation}). This dataset contains the same trajectories as those used in Stage 1. However, unlike Stage 1, we removed the business policies to encourage the model to recall policies within the reasoning (thinking) block. The resulting model is referred to as \ourstwo.

\subsubsection{Stage 3: Reinforcement Learning with GRPO} \label{sec:stage_3}
In the final stage, we employ GRPO \citep{shao_deepseekmath_2024} to reinforce the model's adherence to the required output format and recall policies with high precision. This stage uses 900 unseen assistant turns that were not included in earlier phases. Following \citet{qian_toolrl_2025}, we train the model on every assistant and tool-calling turn. We incorporate five distinct rewards and penalties to guide the model's behavior. We refer to the resulting model as \ours.

\paragraph{Correct Policy Reward:}
We introduce the \textit{PolicyRecall} reward, which rewards generated CoTs to recall the correct and relevant policies while penalizing over-recall. Given the ground-truth policy document and the current generation, an LLM (i.e., \texttt{Qwen3-32B}) first extracts the set of policies ($A$) required to satisfy the user request (see \cref{app:prompt_user_req_to_policy}). We then use the same LLM to extract the set of policies ($B$) referenced in the agent's thinking block (see \cref{app:prompt_cot_to_policy}). However, RL training is highly susceptible to reward hacking \citep{gao2023scaling, tennant2025moralalignmentllmagents, dipta2026ganitllmdifficultyawarebengalimathematical, nazi2026dagdaggerdistractorawaregraphgeneration}, where the model may over-optimize the reward, i.e., by recalling all policies -- thus defeating the intended objective. To mitigate this, we employ the Jaccard score $\mathcal{J}$ as a proxy reward.
\[
    \mathcal{R}_{policy} = \mathcal{J}(A, B) = \frac{|A \cap B|}{
|A \cup B|} \in [0, 1]
\]
The Jaccard score penalizes over-recall while rewarding accurate and relevant policy recall. Specifically, as the model retrieves additional irrelevant policies, the denominator $A \cup B$ increases, causing the score to decrease and thereby discouraging unnecessary recalls.

\paragraph{Hallucinated Policy Penalty:}
Next, we introduce the \textit{HallucinationPenalty}, which penalizes the model whenever a policy is mentioned that does not appear in the ground-truth document. Specifically, we use the same LLM to extract the set of policies ($C$) that are referenced in the model's reasoning but absent from the policy document (see \cref{app:prompt_cot_to_hal_policy}). We then apply the following equation to impose the penalty.
\begin{align}
    \mathcal{P}_{hallucination} = \frac{|C|}{|B| + |C|} \in [0, 1] \nonumber
\end{align}
Here, $B$ and $C$ are disjoint sets of recalled and hallucinated policies, respectively. $|B| + |C|$ equals the total number of mentioned policies.

Both of the above scoring methods, with examples, are illustrated on \cref{fig:policy_recall_reward}. 

\paragraph{Policy Recall Length Penalty:}
In early experiments, we observed that the model often produced excessively long CoTs that repeatedly checked the same policies. We hypothesize that this behavior comes from pretraining on math-reasoning datasets where repeated verification is beneficial; in our setting, however, it only increases cost without improving accuracy \citep{jiang2026rocktokens}.
To mitigate that, inspired by overlong punishment from DAPO \citep{yu_dapo_2025}, we have adopted the same length penalty but only for the CoT block. 
We set $L_{soft} = 100$ and $L_{hard} = 250$.

{\small
\begin{align}
\mathcal{P}_{\text{policy\_len}} &= \begin{cases}
    0, & |y| \leq L_{soft} \nonumber \\
    \frac{(|y| - L_{soft})}{L_{hard} - L_{soft}}, & L_{soft} < |y| \leq L_{hard} \nonumber \\
    1, & L_{hard} < |y| \nonumber
\end{cases}
\end{align}}

\paragraph{Turn Reward:} 
For the tool call turn, following \citet{qian_toolrl_2025}, we use a combination of tool name, parameter name and parameter content matching as the reward score. For the assistant turn, we use \texttt{Qwen3-32B-thinking} to score the generation based on the ground truth response. Details on both of the scores are presented in \cref{app:correctness_reward}.

{\small
\[
\mathcal{R}_{turn} = \begin{cases} 
    \mathcal{RM}(G, P) \in [-3, 3] & \text{if \texttt{assistant}} \\\\
    \mathcal{R}_{tool\_correct} \in [-3, 3] & \text{if \texttt{tool-call}} \\
\end{cases}
\]}

where $\mathcal{RM}(G, P)$ denotes the score assigned by the reward model given the ground-truth response $G$ and the predicted response $P$. The term $\mathcal{R}_{tool\_correct}$ includes $r_{name}$, $r_{param}$, and $r_{value}$ -- the rewards associated with correctly generating the function name, function parameters, and function values, respectively. 

\begin{table}[!t]
\centering
\resizebox{\columnwidth}{!}{
    \begin{tabular}{@{}l|cccc@{}}
    \toprule
    \multirow{2}{*}{\textbf{Agent Model}} & \multicolumn{3}{c}{\textbf{Pass@1}$\uparrow$} & \textbf{Mean} \\
    
     & \textbf{$\boldsymbol{\tau}$-Airline}
     & \textbf{$\boldsymbol{\tau}$-Retail} & \textbf{Avg.} & \textbf{Words$\downarrow$}
     \\
    \midrule
    \multicolumn{5}{c}{\textit{Full Business Policy In-Context}} \\ \midrule
    Claude 3.5 Sonnet &	48.05 & 69.54 & \gca{58.80} & - \\
    
    Claude 3.7 Sonnet & 44.05 & \textbf{80.20} & \gca{62.12} & - \\
    Claude 4 Sonnet & 50.70 & 68.00 & \gca{59.35} & - \\
    GLM-4.5-Air & 54.45 & 75.37 & \gca{64.91} & 42k \\
    GLM-4.5 & \textbf{55.35} & 79.50 & \textbf{\gca{67.42}} & 40.3k \\
    xLAM-2-32b & 37.35 & 63.70 & \gca{50.52} & 45k \\
    Qwen-2.5-32B & 27.95 & 58.04 & \gca{43.00} & 39k \\
    
    \midrule \multicolumn{5}{c}{\textit{Without Business Policy}} \\ \midrule
    xLAM-2-32b & 17.40 & 58.41 & \gca{37.91} & 29.7k \\
    \oursone & 17.15 & 58.72 & \gca{37.93} & 30k \\
    \ourstwo & 36.95 & 63.07 & \gca{50.01} & 45k \\
    \ours & \textbf{42.00} & \textbf{65.51} & \textbf{\gca{53.75}} & \textbf{27k} \\
    \bottomrule
    \end{tabular}
    }
    \caption{\small Success Rate ($pass@1$) and Mean number of words in a trajectory for various proprietary, open-source, and our fine-tuned models on the Retail and Airline domain of \taubench\ (averaged over 5 trials). \textit{Avg.} denotes the average performance across both domains. \oursone, \ourstwo, and \ours correspond to Stage~1, Stage~2, and Stage~3 training, respectively.}
    \label{tab:main_results}
    \vspace{-4mm}
\end{table}

\paragraph{Format Reward:}
Lastly, we use a binary (0/1) reward based on the specified format (see \cref{app:prompt_format_reward}).

Finally we combine all the scores to get the final reward,
\begin{align}
\mathcal{R}_{final} &= \mathcal{R}_{format} + \mathcal{R}_{turn} + \mathcal{R}_{policy} \nonumber \\
&\quad - \mathcal{P}_{hallucination} - \mathcal{P}_{policy\_len} \in [-5, 5] \nonumber
\end{align}
\vspace{-6mm}

\section{Experiment}
\subsection{Training Dataset}

For training, we use the general function-calling dataset \texttt{actionstudio-98k} (GFC) \citep{zhang_actionstudio_2025}, along with the domain-specific tool-calling dataset (APIGen-MT) introduced by \citet{prabhakar_apigen-mt_2025}. APIGen-MT is particularly well-suited for our framework, as it follows the same business policies as our evaluation dataset, \taubench \citep{yao2024tau}. While both our training and test data share the same policy descriptions, during evaluation we don't use those policies.

During analysis, we identified several trajectories in APIGen-MT that contain hallucinated tool calls. After removing these, we retain 4.8k high-quality trajectories. We then randomly sample 50 trajectories from each domain to reserve for Stage 3 GRPO training, referring to this subset as \textit{APIGen-MT-GRPO}. The remaining 4.7k trajectories are used directly for Stage 1 training and are further augmented with synthetically generated CoTs produced by our pipeline (described in \cref{sec:cot_generation}) for Stage 2 training.

\begin{table}[t]
\centering
\small
\begin{tabular}{@{}lccr@{}}
\toprule
\textbf{Agent Model} & \textbf{Input} $\downarrow$ & \textbf{Output} $\downarrow$ & \textbf{Total} $\downarrow$ \\
\midrule
\multicolumn{4}{c}{\textit{With Business Policy}} \\
\midrule
GLM-4.5-Air & 40.3k & 1.5k & 42k \\
GLM-4.5 & 39k & 1.6k & 40.3k \\
xLAM-2-32B & 40k & 5k & 45k \\
Qwen-2.5-32B & 39k & 4k & 43k \\
\midrule
\multicolumn{4}{c}{\textit{Without Business Policy}} \\
\midrule
\oursone & 25k & 5k & 30k \\
\ourstwo & 25k & 20k & 45k \\
\ours \small{\textcolor{gray}{\textit{w/o Length Penalty}}} & 25k & 14k & 39k \\
\ours & 25k & 2.6k & 27k \\
\bottomrule
\end{tabular}
\caption{Mean number of input, output and total words for a single trajectory. Input consists of user request and tool response, while output consists of assistant response and tool call turn.}
\label{tab:compression_ratio}
\vspace{-2mm}
\end{table}

\subsection{Evaluation Dataset \& Metric}
We use \taubench \citep{yao2024tau} as our primary evaluation dataset. We evaluate on both the airline and retail domains. Following \citet{prabhakar_apigen-mt_2025}, we use pass@1 as the main evaluation metric. Detailed implementation information is provided in \cref{app:implementation_details}.

\subsection{Baselines}
We compare our model with several baselines to demonstrate the effect of our method.

\textbf{With Business Policy:} We have used several \textbf{\textit{closed source models:}} (1) Claude-3.5, (2) Claude-3.7, (3) Claude-4; and \textbf{\textit{open source models:}} (4) GLM-4.5, (5) Qwen-2.5, (6) xLAM-2 as baselines.

\textbf{Without Business Policy:} Since the original closed- and open-source models are not trained on the exact policy set, we evaluate three categories of baselines. 
\textbf{\textit{(i) Continual-SFT only:}} base instruction-tuned models further trained on GFC and APIGen-MT (\cref{sec:stage_1}) -- \oursone (Qwen-2.5). 
\textbf{\textit{(ii) Continual-SFT + CoT-SFT:}} models fine-tuned on GFC and APIGen-MT, followed by CoT-augmented APIGen-MT (\cref{sec:stage_2}) -- \ourstwo (Qwen-2.5). 
\textbf{\textit{(iii) Open-source SFT on the same domain:}} open-source SOTA models directly fine-tuned on the APIGen-MT dataset -- xLAM-2 \citep{prabhakar_apigen-mt_2025}.

\textbf{Retrieval and Compression:} we also compare against two families of methods that shorten the policy in context instead of internalizing it, both run on the same base model as \ours (\texttt{Qwen2.5-32B-Instruct}). The first family retrieves policy text at every turn. We use BM25 \citep{robertson2009probabilistic} and a dense retriever \citep{chen-etal-2024-m3}, each at $k \in \{3, 5, 10\}$, over the whole policy document and over the same document split into individual policies. We also add an LLM router that reads the document at every turn and returns the policies it judges relevant. The second family compresses the document instead of selecting from it. We use LLMLingua-2 \citep{pan-etal-2024-llmlingua} and abstractive summaries, each at four length budgets, and the query-aware LongLLMLingua \citep{jiang-etal-2024-longllmlingua}. This gives 47 runs over 26 configurations, all reported in \cref{app:rag_compression_full} (\cref{tab:rag_compression_airline,tab:rag_compression_retail}).

We note that the comparison between our trained models and the zero-shot baselines is not a strict like-for-like evaluation. Instead, these results highlight the value of policy distillation in improving performance and efficiency.

\section{Results \& Analysis}
Our primary results on the \taubench dataset are presented in \cref{tab:main_results}. The findings indicate that while proprietary or extremely large models (e.g., GLM-4.5) achieve higher pass@k scores, they do so at the cost of substantially larger model sizes, increased computational requirements, higher latency, and greater inference cost. In contrast, our models are an order of magnitude smaller than GLM-4.5 (32B vs.\ 355B) and, even without in-context business policies, outperform similarly sized models that rely on such policies.

\begin{table}[!t]
\centering
\small
    \begin{tabular}{@{}lr@{}}
    \toprule
    \textbf{Model} & \textbf{Accuracy (\%)$\uparrow$} \\
    \midrule
    Random & 50.00 \\
    xLAM-2-32b & 55.10 \\
    \oursone & 51.02 \\
    \ourstwo & 69.39 \\
    \ourstwo \small{\textcolor{gray}{\textit{w/o Continual SFT (stage 1)}}} & 67.35 \\
    \ours & \textbf{71.43} \\
    \bottomrule
    \end{tabular}
\caption{QA-based knowledge test result on different models and stages of our method. The generation is sampled using temperature 0.0 for reproducibility.}
\label{tab:knowledge_test}
\vspace{-4mm}
\end{table}

Specifically, our Stage~1 model (\oursone), trained solely via continual fine-tuning, achieves performance comparable to xLAM-2, which is expected given that both models are fine-tuned on similar data. Stage~2, which incorporates CoT-augmented SFT, yields a substantial improvement, boosting pass@1 from $\sim$37\% to $\sim$50\%. However, we observe an increased word count due to the newly introduced thinking block. Finally, Stage~3 with GRPO further improves accuracy while significantly reducing word usage.

We find that most of the gains come from the ``Airline'' domain (\cref{tab:main_results}), whereas improvements in the ``Retail'' domain are considerably smaller (\textbf{141\% relative improvement over xLAM in the Airline domain vs. 12\% in Retail}). This observation is consistent with the findings of \citet{yao2024tau}, which reports only minor performance drops in the ``Retail'' setting when business policies are removed. 
Through a fine-grained examination of the ground-truth policy documents, we find that the ``airline'' policies are substantially more out-of-distribution, unlikely to have appeared during pre-training. As a result, pretrained models have little to no parametric knowledge of this domain. In contrast, the ``retail'' policies are very common across many retail-oriented datasets, making it far more likely that pretrained models already have those in their parametric knowledge. A more detailed analysis with examples of different domains is shown in \cref{app:airline_vs_retail}.

\begin{figure}
    \centering
    \includegraphics[width=1\linewidth]{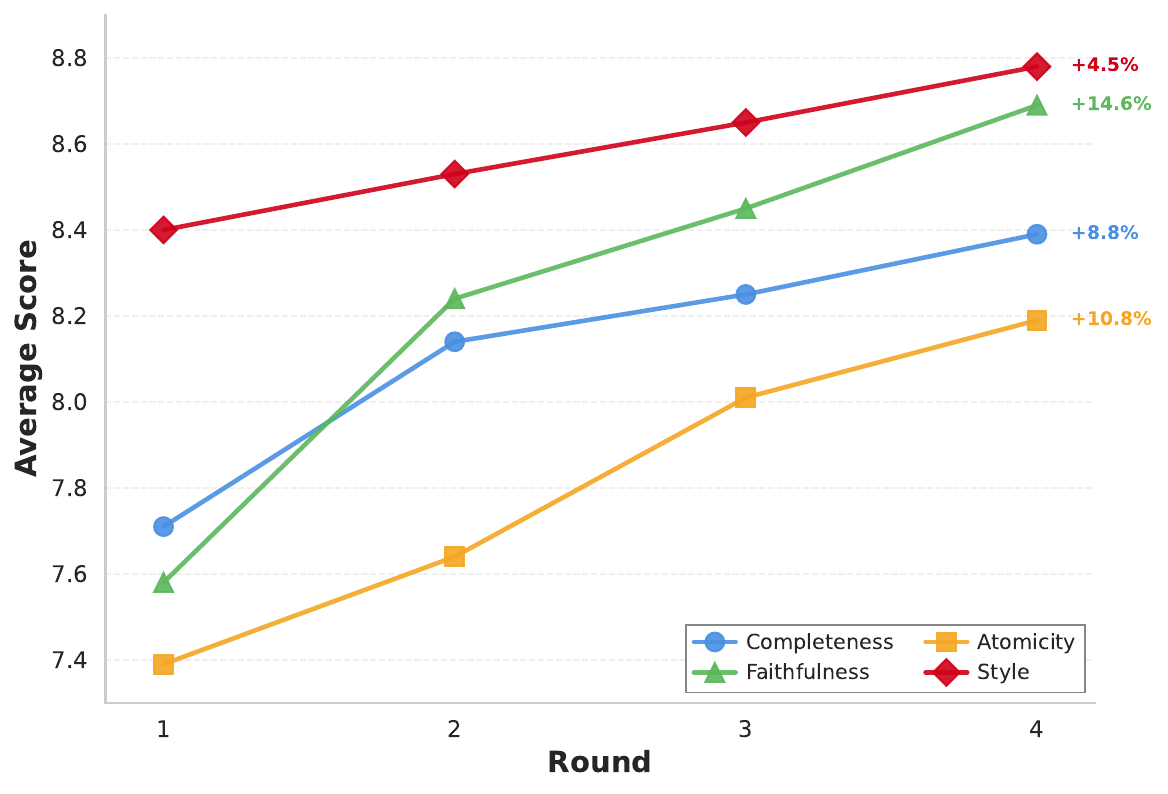}
    \caption{Evolution of CoT quality metrics through iterative generation-refinement, showing consistent improvements across all dimensions (4.5\%--14.6\% gains).}

    \label{fig:cot_analysis}
\end{figure}

\subsection{Compression Ratio}
To further validate the compression ratio achieved by our method, we provide a fine-grained analysis of both input and output word counts in \cref{tab:compression_ratio}. Here, the input includes the user request and the tool response returned from the tool call, while the output includes both the assistant response and the generated tool call. We report word counts rather than tokens to ensure a fair comparison across models with different tokenizers. The results show that our final-stage model produces the smallest total word count among all models and configurations.

A deeper analysis reveals that our Stage~1 model (\oursone), trained only with continual SFT, generates a comparable number of words to its counterparts. In Stage~2 (\ourstwo), however, the introduction of CoT reasoning substantially increases output length. Because the thinking block resembles behavior in mathematical reasoning tasks at this stage, the model often re-evaluates the same policy multiple times for verification, leading to a significant increase in output words. Even \ours, when trained without the length penalty, does not meaningfully reduce output length -- highlighting the necessity of the length penalty. With all rewards and penalties included, our final model demonstrates the most efficient token usage.

We emphasize that the token savings reported above is for inference time. Although our method has an upfront training cost (see \cref{app:implementation_details}), this one-time expense is offset by the reduced cost of serving thousands of queries daily.

\subsection{Retrieval and Compression Baselines}
\begin{table}[t]
\centering
\resizebox{\columnwidth}{!}{
\begin{tabular}{@{}l|cccc@{}}
\toprule
\multirow{2}{*}{\textbf{Policy Handling}} & \multicolumn{3}{c}{\textbf{Pass@1}$\uparrow$} & \textbf{Mean} \\
 & \textbf{$\boldsymbol{\tau}$-Airline} & \textbf{$\boldsymbol{\tau}$-Retail} & \textbf{Avg.} & \textbf{Words}$\downarrow$ \\
\midrule
No policy & 15.60 & 48.70 & \gca{32.15} & 29.0k \\
\midrule \multicolumn{5}{c}{\textit{Retrieval}} \\ \midrule
BM25, $k{=}5$ & 22.80 & 54.09 & \gca{38.45} & 34.5k \\
Dense (atomic), $k{=}5$ & 19.60 & 51.13 & \gca{35.37} & 31.2k \\
LLM router & 25.60 & 49.57 & \gca{37.59} & 57.3k \\
\midrule \multicolumn{5}{c}{\textit{Compression}} \\ \midrule
Extractive & 15.60 & 54.26 & \gca{34.93} & 37.2k \\
Abstractive summary & 28.00 & 54.96 & \gca{41.48} & 42.5k \\
Query-aware & 14.00 & 49.22 & \gca{31.61} & 47.5k \\
\midrule
\ours & \textbf{42.00} & \textbf{65.51} & \textbf{\gca{53.75}} & \textbf{27k} \\
\bottomrule
\end{tabular}
}
\caption{\small Retrieval and compression baselines, all run on \texttt{Qwen2.5-32B-Instruct} at 5 trials on the full task set with the same user model. Each row is the strongest configuration of its family, selected by mean pass@1 over the two domains; every configuration we ran is listed in \cref{tab:rag_compression_airline,tab:rag_compression_retail}.}
\label{tab:rag_compression}
\end{table}

\cref{tab:rag_compression} reports the strongest configuration of each baseline family, and \cref{app:rag_compression_full} reports all 47. No configuration matches \ours on accuracy. Thirteen airline configurations send fewer words per trajectory, but each of them loses at least 16.4 pass@1 points there. The strongest retriever we tried, BM25 at $k{=}5$, gives up 15.30 pass@1 points, and the best compressor, an abstractive summary, gives up 12.27 points while sending 15.5k more words per trajectory. The LLM router is the most expensive configuration in the grid, because it re-reads the whole policy document at every turn, and it still loses 16.16 points.

\subsection{QA-based Knowledge Test}
To assess whether the trained models have acquired business policy knowledge,
we manually create 49 yes/no-questions from the \taubench airline domain. These questions were manually curated by the authors from the policy document. The results of the knowledge test are provided in \cref{tab:knowledge_test}. We have used a simple QA prompt to ask the question.

As expected, the primary improvement in policy knowledge occurs during the CoT-augmented SFT stage. Continual-SFT alone does not meaningfully improve policy recall.
In contrast, CoT training explicitly forces the model to recall and apply the relevant policies. The final GRPO stage primarily contributes to reducing hallucinations (through the PolicyRecall penalty) and decreasing output length (through the Length Penalty).

\subsection{Ablation Study}

\subsubsection*{Impact of CoT Filtering} \label{app:cot_analysis}
\begin{table}[t]
\centering
\small
\begin{tabular}{@{}lcccc@{}}
\toprule
\multirow{2}{*}{\textbf{Model}} & \textbf{Filtered} & \multicolumn{3}{c}{\textbf{Pass@1} $\uparrow$} \\
 & \textbf{CoT?} & \textbf{Airline} & \textbf{Retail} & \textbf{Avg.} \\
\midrule
\ourstwo & \texttimes & 36.85 & 61.22 & 49.04 \\
\ourstwo & \checkmark & \textbf{36.95} & \textbf{63.07} & \textbf{50.01} \\
\midrule
\ours & \texttimes & 37.40 & 60.20 & 48.80 \\
\ours & \checkmark & \textbf{42.00} & \textbf{65.51} & \textbf{53.75} \\
\bottomrule
\end{tabular}
\caption{Performance comparison of filtered vs. unfiltered Chain-of-Thought training.}
\label{tab:impact_of_cot_filtering}
\vspace{-2mm}
\end{table}

Recall that we employ a multi-stage filtering–refinement pipeline to generate CoTs for each turn, which introduces a significant computational cost. While a more cost-effective approach would be to use the CoTs from the first round directly, we examine the necessity of multi-stage filtering and refinement from quantitative, qualitative, and end-performance perspectives. 

First, \cref{fig:cot_analysis} illustrates the improvement in rubric scores across refinement stages. The results show a clear positive impact of multi-stage filtering on the overall quality of the generated CoTs. The most notable gains occur in \textit{Faithfulness}, which is the most crucial metric for a policy-adherent agents. We also observe substantial improvements in both \textit{Completeness} and \textit{Atomicity}, further motivating the need for iterative refinement. Overall, the average rubric score increases from $\sim$7.8 to $\sim$8.5.

\begin{figure*}[!t]
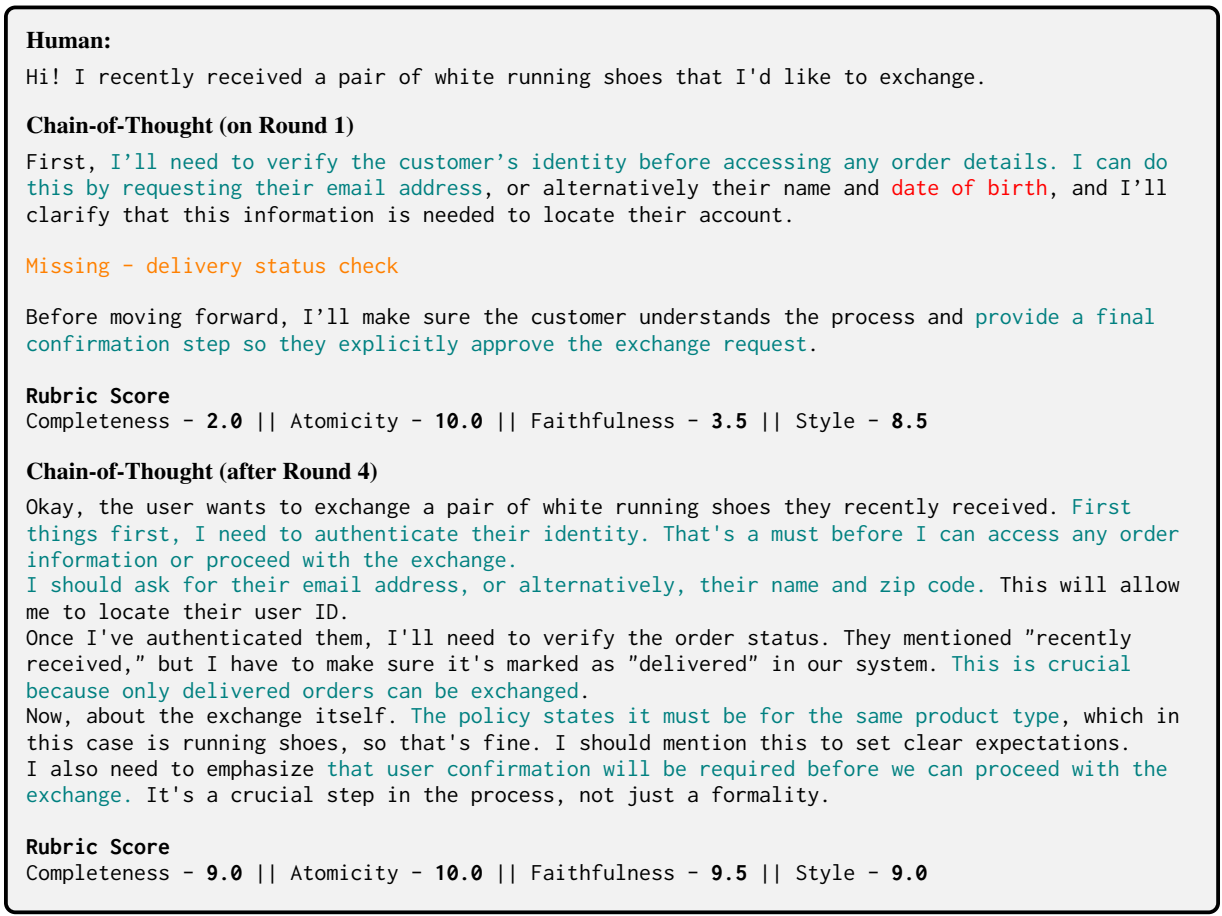

    \centering
\begin{tcolorbox}[
    coltitle=white,
    colframe=black,
    colback=black!5!white,
    breakable,
    enhanced,
    fontupper=\footnotesize,
    fontlower=\footnotesize,
    fonttitle=\small,
    left=4pt,
    right=4pt
]
\textbf{Human:}
\vspace{-2mm}
\begin{Verbatim}[breaklines=true,breaksymbol=,fontfamily=tt,commandchars=\\\{\}]
Hi! I recently received a pair of white running shoes that I'd like to exchange.
\end{Verbatim}
\textbf{\small Chain-of-Thought (on Round 1)}
\vspace{-2mm}
\begin{Verbatim}[breaklines=true,breaksymbol=,fontfamily=tt,commandchars=\\\{\}]
First, {\color{teal}I’ll need to verify the customer’s identity before accessing any order details. I can do this by requesting their email address}, or alternatively their name and {\color{red}date of birth}, and I’ll clarify that this information is needed to locate their account.

{\color{orange}Missing - delivery status check}

Before moving forward, I’ll make sure the customer understands the process and {\color{teal}provide a final confirmation step so they explicitly approve the exchange request}.

\textbf{Rubric Score}
Completeness - \textbf{2.0} || Atomicity - \textbf{10.0} || Faithfulness - \textbf{3.5} || Style - \textbf{8.5}
\end{Verbatim}
\textbf{\small Chain-of-Thought (after Round 4)}
\vspace{-2mm}
\begin{Verbatim}[breaklines=true,breaksymbol=,fontfamily=tt,commandchars=\\\{\}]
Okay, the user wants to exchange a pair of white running shoes they recently received. {\color{teal}First things first, I need to authenticate their identity. That's a must before I can access any order information or proceed with the exchange.}
{\color{teal}I should ask for their email address, or alternatively, their name and zip code.} This will allow me to locate their user ID. 
Once I've authenticated them, I'll need to verify the order status. They mentioned "recently received," but I have to make sure it's marked as "delivered" in our system. {\color{teal}This is crucial because only delivered orders can be exchanged}.
Now, about the exchange itself. {\color{teal}The policy states it must be for the same product type}, which in this case is running shoes, so that's fine. I should mention this to set clear expectations.
I also need to emphasize {\color{teal}that user confirmation will be required before we can proceed with the exchange.} It's a crucial step in the process, not just a formality.

\textbf{Rubric Score}
Completeness - \textbf{9.0} || Atomicity - \textbf{10.0} || Faithfulness - \textbf{9.5} || Style - \textbf{9.0}
\end{Verbatim}
\end{tcolorbox}
    \vspace{-3mm}
    \caption{Example CoTs across different rounds of filtering and refinement. Correct and relevant policies are shown in {\color{teal}green}, hallucinated policies in {\color{red}red}, and missing policies in {\color{orange}orange}.}
    \label{fig:cot_qualitative_analysis}
    \vspace{-3mm}
\end{figure*}

Next, from a qualitative standpoint, in \cref{fig:cot_qualitative_analysis}, we compare two CoTs generated at different stages. As shown, the initial round typically retrieves some relevant policies but often misses crucial ones and occasionally hallucinates information. By evaluating these CoTs and providing refinement summaries, the subsequent stages are able to steer the reasoning in the correct direction, producing more accurate and policy-grounded CoTs.

Finally, from an end-performance perspective, we examine the impact of CoT filtering in \cref{tab:impact_of_cot_filtering}. 
The results show that filtered CoTs provide a clear benefit for GRPO-based training (48.80 $\rightarrow$ 53.75), while the effect on SFT alone is more modest (49.04 $\rightarrow$ 50.01). This suggests that CoT quality matters most when the training objective explicitly optimizes for policy recall.
We hypothesize that using incomplete or hallucinated CoTs during Stage~2 introduces errors that may push the model into a suboptimal region from which recovery is difficult. Even though the filtered CoTs do not directly influence the Stage~3 training data, the GRPO stage begins from the Stage~2 checkpoint; therefore, noisy or low-quality CoTs in Stage~2 can negatively shift the model's distribution, ultimately harming final performance.

In practice, the extra compute cost of multi-round filtering is outweighed by the substantial gains in both CoT quality (\cref{fig:cot_analysis,fig:cot_qualitative_analysis}) and downstream performance (\cref{tab:impact_of_cot_filtering}).

\begin{table}[!t]
\centering
\small
\begin{tabular}{@{}lccc@{}}
\toprule
\multirow{2}{*}{\textbf{Model}} & \multicolumn{3}{c}{\textbf{Pass@1} $\uparrow$} \\
 & \textbf{Airline} & \textbf{Retail} & \textbf{Avg.} \\
\midrule
\ourstwo + TF & 40.40 & 56.52 & 48.46 \\
\ourstwo + TF + PR & 38.00 & 60.52 & 49.26 \\
\ourstwo + TF + PR + HP & 41.33 & 62.61 & 51.97 \\
\ourstwo + TF + PR + HP + LP & \textbf{42.00} & \textbf{65.51} & \textbf{53.75} \\
\bottomrule
\end{tabular}
\caption{Model performance with various reward scores. TF = Turn, Format reward, PR = correct policy-recall reward, HP = hallucination penalty, LP = length penalty.}
\label{tab:impact_of_policy_recall}
\vspace{-4mm}
\end{table}

\subsection*{Impact of Policy Recall Reward}
In this experiment, we ablate the PolicyRecall rewards introduced in \cref{sec:multi_stage_training}. Recall that the PolicyRecall reward includes three components: Correct Policy Reward, Hallucination Penalty, and Policy Recall Length Penalty. In \cref{tab:impact_of_policy_recall}, we incrementally add each component to our Stage~2 model to examine its contribution to final performance.

We first observe that adding \textbf{only} the Correct Policy Reward actually degrades performance in the Airline domain (40.40 $\rightarrow$ 38.00). The Airline domain is more out-of-distribution relative to the model’s pre-training distribution (see \cref{app:airline_vs_retail} for further discussion). We hypothesize that introducing only the Correct Policy reward increases the risk of hallucination in this more OOD setting.

Adding a Hallucination Penalty mitigates this issue, leading to improvements in Pass@1 across both domains. Finally, incorporating a Length Penalty reduces token usage while further improving performance.
These results highlight the importance of both the PolicyRecall reward and the Hallucination Penalty in improving accuracy, while the Length Penalty plays a crucial role in reducing overall token cost (\cref{tab:compression_ratio}).
An ablation of overriding already learned policies is provided in \cref{app:override_policy}.

\subsection{Error Analysis}
We discuss three sources of error.

\textbf{Hallucinated and missing policies:} within a single reasoning trace, the two recurring errors are policies the model invents and policies it omits. \cref{fig:cot_qualitative_analysis} shows both in the same round-1 trace, where the model asks for a date of birth that no policy mentions and skips the delivery-status check the request needs. Refinement fixes both, and the largest rubric gain in \cref{fig:cot_analysis} is on \textit{Faithfulness}, which penalizes hallucinated policies.

\textbf{Conditional rules in the Airline domain:} most errors that survive training sit in the Airline domain, where pass@1 reaches 42.00 against 65.51 on Retail (\cref{tab:main_results}). The policies involved are the conditional ones: membership-dependent baggage allowances and cancellation rules that depend on fare class and travel insurance (examples in \cref{app:airline_vs_retail}). Retail policies are generic enough that the base model already knows most of them, which is also why the relative gain there is 12\% against 141\% on Airline.

\textbf{Overridden policies:} when a learned policy is contradicted by a new policy given in context, our final model outperforms the SFT baselines but stays only slightly above random (\cref{app:override_policy}). Internalizing a policy set therefore does not make the model defer to an updated one \citep{hossain2026rightknowledgewronganswer}, which is the failure mode that matters most when policies change.

\section{Conclusion \& Future Work}

Building agents that adhere to business policies without relying on in-context policy descriptions
is a critical challenge. This paper addresses this issue through a multi-stage training strategy %
with a central emphasis on policy recall during CoT reasoning.
We make three key contributions: (1) a framework for automatically extracting high-quality CoT traces that is adaptable across diverse business domains; (2) an RL reward design that combines policy recall rewards with hallucination penalties; and (3) empirical evidence demonstrating substantial performance improvements.
Our findings reveal that SFT improves policy recall on familiar domains but is prone to memorization; RL improves task performance over more OOD data, though whether this constitutes true generalization beyond the evaluated domains remains an open question for future work.
We hope our work--automatic CoT extraction, policy-recall rewards, and systematic SFT-RL evaluation--provides foundational tools for building policy-aware language agents.

\section*{Limitations}
While our method demonstrates strong policy adherence without requiring in-context business policies, it incurs a substantial upfront cost for generating high-quality and reliable chain-of-thought. Owing to the nature of LLMs and LLM-as-a-judge evaluation, there remains a risk of hallucinated policies, especially in rare or under-specified cases. We mitigate this through multi-round evaluation and refinement. In addition, the approach may inherit biases from the underlying LLMs, which can influence both the generated CoTs and the policy judgments. Our rubric scoring and policy extraction each depend on a single judge model, \texttt{Claude-3.5-v1} for the rubrics and \texttt{Qwen3-32B} for the policies. We hold both fixed across every experiment and ablation, so the comparisons we report are internally consistent, although absolute rubric scores may shift under a different judge. The reward also matches policies as sets, so it can miss cases where two policies are worded similarly or where a rule applies only under an implicit condition.

Next, while our method reduces the number of input tokens, it increases the number of output tokens by introducing explicit reasoning. For reasoning models, this increase is not substantial, as they typically produce a reasoning block regardless. However, for non-reasoning models, this may result in a noticeable increase in output token usage. Future work could compress these reasoning traces into latent tokens, following recent approaches \citep{cheng_compressed_2024}, although this direction is beyond the scope of this work.

Our evidence covers two \taubench domains, and our method assumes business policies exist as a machine-readable document. Where no such document exists, the policies must first be written out for our pipeline to apply.

Finally, training on a fixed set of policies risks overfitting to that specific version, reducing reliability when policies change and requiring frequent retraining. One potential solution is to incorporate additional synthetic policy-override data during training to teach the model how to adapt to overridden policies and mitigate overfitting. We leave this direction for future work.

\bibliography{acl,custom}

\appendix
\crefalias{section}{appendix}
\crefalias{subsection}{appendix}
\clearpage

\section*{Appendix}

\section{CoT Evaluation Thresholds}
\label{app:thresholds}

Table~\ref{tab:thresholds} reports the rubric thresholds used in the 
evaluation--filtering cycle (\cref{sec:cot_generation}). A CoT must meet 
\emph{all} thresholds simultaneously to pass; otherwise it is sent to 
the refinement stage. The thresholds reflect our design priority: 
Faithfulness is the strictest (a perfect score of 10 is required, as 
any hallucinated policy is unacceptable), followed by Completeness, 
Atomicity, and Style. These values were hand-engineered based on 
qualitative inspection of CoTs near each boundary and kept fixed 
throughout all experiments.

\begin{table}[h]
\centering
\small
\begin{tabular}{lc}
\toprule
\textbf{Rubric} & \textbf{Threshold ($\geq$)} \\
\midrule
Faithfulness   & 10 \\
Completeness   & 9  \\
Atomicity      & 7  \\
Style          & 6  \\
\bottomrule
\end{tabular}
\caption{Minimum rubric scores (on a 1--10 scale) required for a 
generated CoT to pass the evaluation stage.}
\label{tab:thresholds}
\end{table}

\begin{table*}[t]
\centering
\small
\begin{tabular}{clcll}
\toprule
\textbf{Stage} & \textbf{Data} & \textbf{Policy In-Context?} & \textbf{Objective} & \textbf{Model} \\
\midrule
1 & GFC + APIGen-MT & \checkmark & General + Policy tool-calling & \oursone \\
2 & CoT-augmented APIGen-MT & \texttimes & Policy Recall via CoT & \ourstwo \\
3 & APIGen-MT-GRPO (900 unseen) & \texttimes & RL for Precise Policy Recall & \ours \\
\bottomrule
\end{tabular}
\caption{Overview of the three-stage training recipe. Each stage builds on the checkpoint from the previous stage.}
\label{tab:training-stages}
\end{table*}

\section{Airline vs. Retail Policy} \label{app:airline_vs_retail}
The results in \cref{tab:main_results} shows that the improvement gain in airline domain is much higher than the retail domain. Through manual analysis, we identify the following reasons.

\subsection{Uncommon policies}
These policies differ numerically or conceptually from those found in real-world datasets. For example, ``each reservation can have at most \textbf{five} passengers,'' whereas in most real-world scenarios, airlines allow an arbitrary number of passengers. Additional examples are provided below:

\begin{itemize}[left=2pt,itemsep=2pt]
    \item Each reservation can use at most one travel certificate, at most one credit card, and at most three gift cards.
    \item  The remaining amount of a travel certificate is not refundable.
    \item The user can add but not remove checked bags. 
    \item The user cannot add insurance after initial booking.
    \item The user can modify passengers but cannot modify the number of passengers. This is something that even a human agent cannot assist with.
    \item basic economy or economy flights can be cancelled only if travel insurance is bought and the condition is met, and business flights can always be cancelled. The rules are strict regardless of the membership status. The API does not check these for the agent, so the agent must make sure the rules apply before calling the API!
\end{itemize}

\subsection{Conditional Rules}
These rules depend on conditions that are highly specific to a particular business context. Additional examples are provided below:

\begin{itemize}[left=2pt,itemsep=2pt]
    \item If the booking user is a regular member, 0 free checked bag for each basic economy passenger, 1 free checked bag for each economy passenger, and 2 free checked bags for each business passenger.
    \item If the booking user is a silver member, 1 free checked bag for each basic economy passenger, 2 free checked bag for each economy passenger, and 3 free checked bags for each business passenger.
    \item If the booking user is a gold member, 2 free checked bag for each basic economy passenger, 3 free checked bag for each economy passenger, and 3 free checked bags for each business passenger.
    \item Each extra baggage is 50 dollars.
    \item If the user is silver/gold member or has travel insurance or flies business, and complains about cancelled flights in a reservation, the agent can offer a certificate as a gesture after confirming the facts, with the amount being \$100 times the number of passengers.
    \item Do not compensate if the user is regular member and has no travel insurance and flies (basic) economy.
\end{itemize}

\subsection{What about Retail?}
The retail policies are generally generic, such as ``returns will arrive in 5--7 business days'' or ``products must be returned within 30 days.'' 
Upon closer inspection, we identified only a single policy that is truly specific to the business:

\begin{itemize}[left=2pt,itemsep=2pt]
\item Our retail store has 50 types of products. For each type of product, there are variant items of different options. For example, for a 't shirt' product, there could be an item with option 'color blue size M', and another item with option 'color red size L'.
\end{itemize}

As illustrated, the number of uncommon or highly specific policies is substantially higher in the Airline domain. In contrast, the Retail domain contains largely generic rules (e.g., standard return windows) that are common across most retail businesses. While this reflects a limitation of the \taubench dataset, real-world business policies tend to be far more domain-specific. \textbf{This also explains the smaller performance gains observed in the Retail setting compared to the Airline domain}.

\section{Override Policies} \label{app:override_policy}
\begin{figure*}[!t]
    \centering
    \begin{minipage}[b]{0.48\textwidth}
        \centering
        \includegraphics[width=0.99\columnwidth]{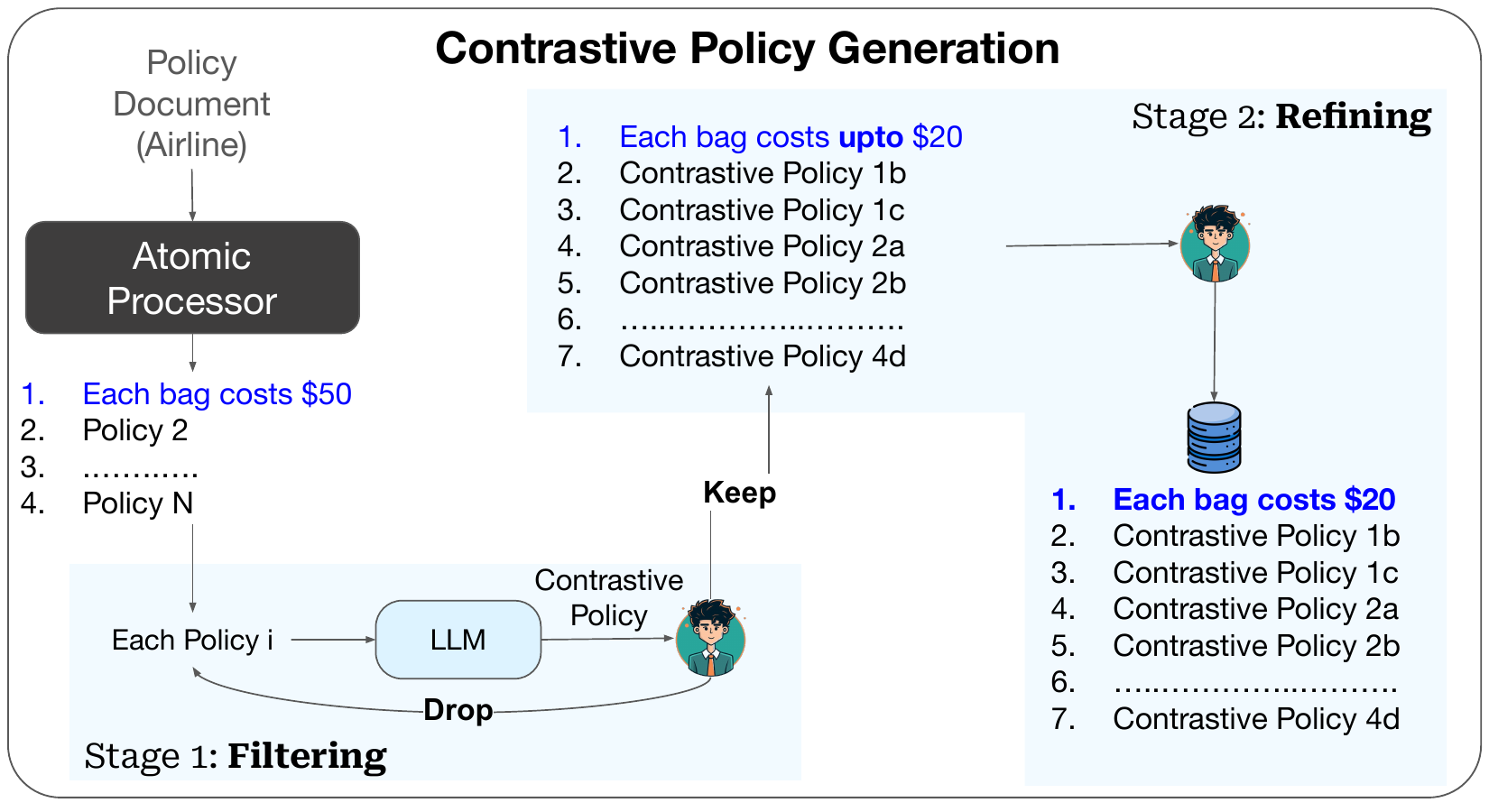}
    \end{minipage}
    \begin{minipage}[b]{0.48\textwidth}
        \centering
        \includegraphics[width=0.99\columnwidth]{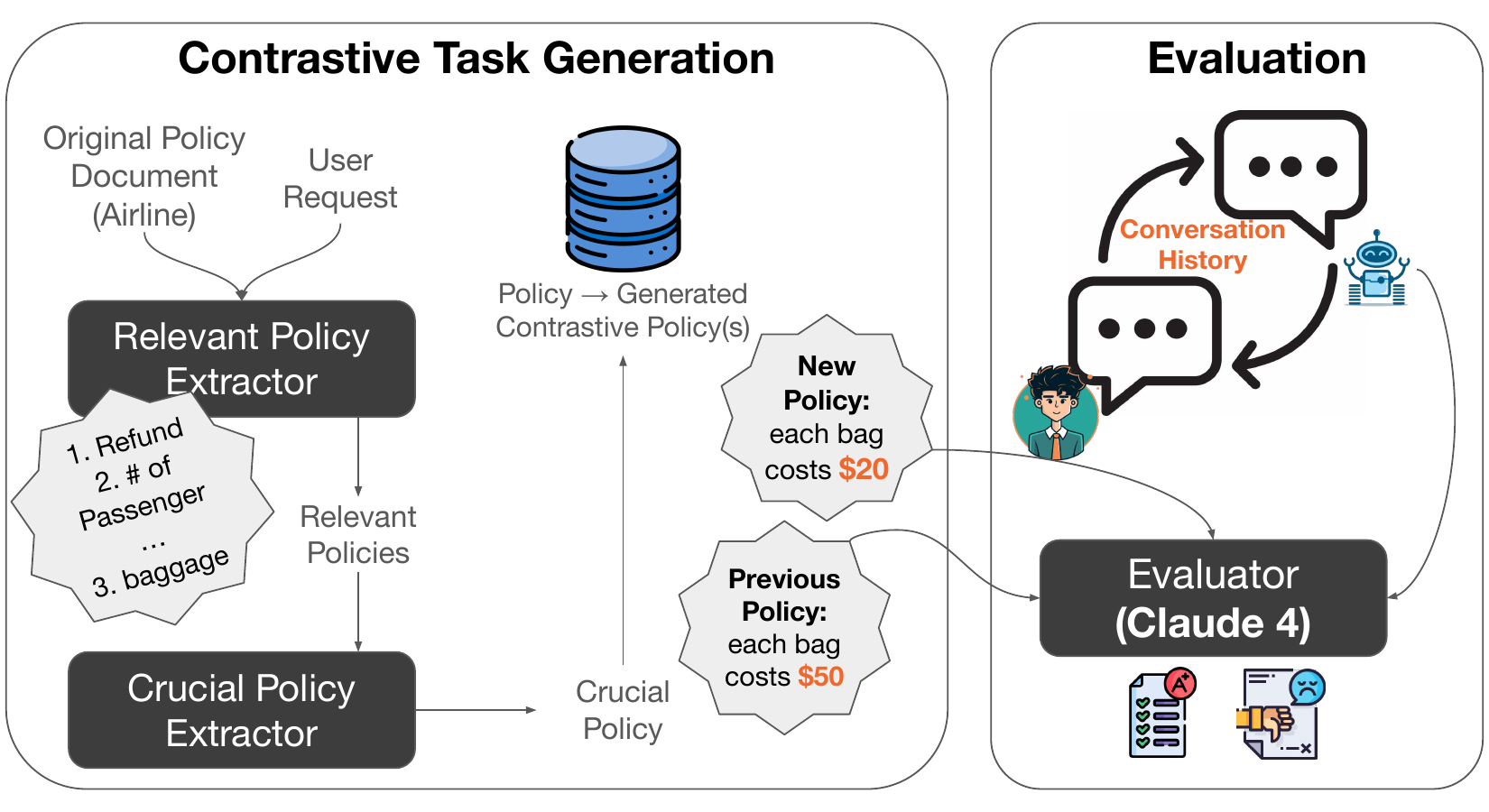}
    \end{minipage}

    \caption{Overview of the Override Policy task generation and evaluation pipeline, consisting of Contrastive Policy Generation, Contrastive Task Generation, and Policy Overriding Evaluation.}

    \label{fig:override_policy}
\end{figure*}

Business policies often change frequently--sometimes even daily--making it impractical to retrain a model each time a policy is updated. Therefore, in this section, we will explore if we can override some of the already learned business policies through in-context prompting. 

To evaluate this capability, we synthetically augment the \taubench dataset. Given the OOD nature of the ``Airline'' domain (details in \cref{app:airline_vs_retail}), we built $\tau$-overrule on top of the airline domain. 
\cref{fig:override_policy} summarizes our pipeline for constructing override tasks by replacing a single crucial policy with its contrastive counterpart and then checking whether the agent follows the new policy in context.

\subsection{Contrastive Policy Generation}
We generate contrastive policies that are easily verifiable (left box in \cref{fig:override_policy}). This process employs a human-in-the-loop system. Specifically, for each atomic policy in the airline domain, we prompt an LLM to generate up to 10 contrastive policies that contradict the original policy. Each generated policy is then reviewed and filtered by the authors and labeled as either \textit{keep} or \textit{drop}.

Following this initial generation-filtering loop, all \textit{kept} contrastive policies undergo a second round of human refinement. In this stage, the goal is to revise the policies to ensure they are both clearly written and directly oppose the original policy.

\subsection{Contrastive Task Generation}
At this stage (middle box in \cref{fig:override_policy}), we use the previously generated contrastive policies to create override tasks. \taubench includes 50 tasks in the airline domain. For each task, we first use an LLM-based extractor to identify all relevant policies necessary to fulfill the user request.

Our initial analysis revealed that not all policies are suitable for replacement, particularly those with trivial impact, i.e., a policy like ``ask for explicit yes before running any database call'' is difficult to validate if overridden with ``don’t ask for explicit yes,'' as it minimally affects task outcome and is hard to verify in conversation.

To address this, we use another LLM to extract the most critical policy from all the relevant policies--one whose alteration would make it hard to satisfy the user request. We then replace this crucial policy with its contrastive counterpart from our previously constructed contrastive policy database.

\subsection{Evaluation}
During evaluation, we first let the agent model and a user-simulated LLM to generate the full conversation given the override policy in context (see \cref{app:prompt_override_prompt} for the exact prompt). Next, we employ a powerful LLM-as-a-judge (i.e. \texttt{Claude-3.7-Sonnet}) to assess whether the agent adhered to the overridden policy.

\subsection{Results}
\begin{table}[h]
\small
\centering
\begin{tabular}{@{}lr@{}}
\toprule
\textbf{Agent Model} & \textbf{Override Accuracy$\uparrow$} \\
\midrule
Random & 50.00 \\
xLAM-2-32b & 33.87 \\
\oursone & 37.87 \\
\ourstwo & 47.01 \\
\ours & \textbf{53.33} \\
\bottomrule
\end{tabular}
\caption{Override accuracy of different models in our synthetically generated dataset.}
\label{tab:override_policy}
\end{table}

The results for policy overriding are presented in \cref{tab:override_policy}. While our final model outperforms the SFT baselines on this task, its accuracy remains only slightly above random. We believe that incorporating instruction-following datasets during SFT (Stage~2) and GRPO training (Stage~3) would substantially improve the model's ability to override policies when required. We leave this direction for future work.

\section{Retrieval and Compression Configurations} \label{app:rag_compression_full}
We ran 26 configurations for 47 runs in total, all 26 on the airline domain and 21 of them also on retail, each at 5 trials on the full task set. \texttt{atomic-*} configurations retrieve over atomic policies rather than the whole document, so they draw on a different corpus; four of them were run on airline only. \texttt{basic} is a lower-bound reference that sends five generic rules and no domain policy. The \texttt{full} configuration re-runs the in-context setting reported for \texttt{Qwen2.5-32B-Instruct} in \cref{tab:main_results} on this harness, so it is a separate run and its per-domain scores disagree with \cref{tab:main_results} by about four points in either direction (airline 32.00 vs.\ 27.95, retail 54.09 vs.\ 58.04). Both runs support the same conclusion: keeping the full policy in context averages 43.0 pass@1, far below \ours\ at 53.75.

\begin{table*}[t]
\centering
\begin{tabular}{@{}lrrrr@{}}
\toprule
\textbf{Configuration} & \textbf{Pol./call} & \textbf{Pass@1}$\uparrow$ & \textbf{Pass@5}$\uparrow$ & \textbf{Words}$\downarrow$ \\
\midrule
\texttt{full} & 1051 & 32.00 & 12.00 & 37.3k \\
\texttt{summary-500w}$^{\dagger}$ & 565 & 28.00 & 8.00 & 32.0k \\
\texttt{llmrouter} & 137 + 1540 & 25.60 & 4.00 & 50.6k \\
\texttt{summary-50w}$^{\dagger}$ & 259 & 25.60 & 6.00 & 26.0k \\
\texttt{bm25-k10} & 444 & 24.00 & 8.00 & 30.3k \\
\texttt{summary-250w}$^{\dagger}$ & 469 & 23.60 & 6.00 & 30.4k \\
\texttt{atomic-dense-k10} & 217 & 23.20 & 6.00 & 25.3k \\
\texttt{bm25-k3} & 158 & 22.80 & 4.00 & 23.6k \\
\texttt{bm25-k5} & 245 & 22.80 & 4.00 & 28.3k \\
\texttt{dense-k5} & 257 & 22.40 & 6.00 & 29.0k \\
\texttt{dense-k3} & 165 & 22.00 & 2.00 & 25.9k \\
\texttt{dense-k10} & 471 & 21.20 & 6.00 & 31.1k \\
\texttt{atomic-dense-k5} & 124 & 19.60 & 4.00 & 24.7k \\
\texttt{atomic-bm25-k10} & 198 & 18.80 & 6.00 & 25.1k \\
\texttt{summary-150w}$^{\dagger}$ & 418 & 18.80 & 2.00 & 28.6k \\
\texttt{atomic-llmrouter} & 108 + 1402 & 18.00 & 4.00 & 41.2k \\
\texttt{atomic-bm25-k3} & 78 & 17.60 & 6.00 & 22.1k \\
\texttt{atomic-bm25-k5} & 113 & 17.60 & 6.00 & 23.8k \\
\texttt{basic} & 164 & 16.80 & 4.00 & 28.1k \\
\texttt{atomic-dense-k3} & 85 & 16.40 & 2.00 & 23.6k \\
\texttt{llmlingua2-500w} & 497 & 15.60 & 6.00 & 28.8k \\
\texttt{none} & 0 & 15.60 & 8.00 & 20.3k \\
\texttt{llmlingua2-150w} & 148 & 14.80 & 8.00 & 24.1k \\
\texttt{longllmlingua-250w} & 228 + 1051 & 14.00 & 2.00 & 38.4k \\
\texttt{llmlingua2-50w} & 39 & 12.80 & 2.00 & 22.2k \\
\texttt{llmlingua2-250w} & 242 & 12.00 & 0.00 & 26.9k \\
\bottomrule
\end{tabular}
\caption{All policy-handling configurations on the \textbf{airline} domain, ordered by pass@1. \textit{Pol./call} is the mean number of policy words sent per model call, plus the words the retriever or compressor itself reads where that applies. \textit{Words} is the mean total words per trajectory. $^{\dagger}$the \texttt{-Nw} suffix is the word budget we requested. These rows exceeded it, and \textit{Pol./call} is the length they actually sent.}
\label{tab:rag_compression_airline}
\end{table*}

\begin{table*}[t]
\centering
\begin{tabular}{@{}lrrrr@{}}
\toprule
\textbf{Configuration} & \textbf{Pol./call} & \textbf{Pass@1}$\uparrow$ & \textbf{Pass@5}$\uparrow$ & \textbf{Words}$\downarrow$ \\
\midrule
\texttt{summary-500w}$^{\dagger}$ & 962 & 54.96 & 25.22 & 53.0k \\
\texttt{llmlingua2-500w}$^{\dagger}$ & 508 & 54.26 & 20.87 & 45.5k \\
\texttt{bm25-k5} & 208 & 54.09 & 24.35 & 40.6k \\
\texttt{full} & 1020 & 54.09 & 18.26 & 53.4k \\
\texttt{dense-k10} & 371 & 53.91 & 20.87 & 42.2k \\
\texttt{bm25-k3} & 141 & 53.74 & 19.13 & 38.7k \\
\texttt{summary-50w}$^{\dagger}$ & 413 & 53.39 & 21.74 & 44.6k \\
\texttt{bm25-k10} & 348 & 52.87 & 27.83 & 41.9k \\
\texttt{llmlingua2-250w} & 250 & 52.70 & 20.87 & 40.9k \\
\texttt{summary-250w}$^{\dagger}$ & 506 & 52.70 & 21.74 & 44.9k \\
\texttt{dense-k5} & 206 & 52.17 & 20.00 & 38.8k \\
\texttt{atomic-llmrouter} & 96 + 1091 & 51.83 & 20.00 & 56.2k \\
\texttt{summary-150w}$^{\dagger}$ & 361 & 51.83 & 23.48 & 44.8k \\
\texttt{atomic-dense-k5} & 101 & 51.13 & 20.87 & 37.8k \\
\texttt{atomic-bm25-k5} & 99 & 50.78 & 19.13 & 38.5k \\
\texttt{dense-k3} & 139 & 50.09 & 21.74 & 38.1k \\
\texttt{llmrouter} & 142 + 1524 & 49.57 & 20.87 & 64.0k \\
\texttt{longllmlingua-250w}$^{\dagger}$ & 339 + 1020 & 49.22 & 17.39 & 56.5k \\
\texttt{llmlingua2-50w} & 40 & 49.04 & 20.00 & 38.7k \\
\texttt{none} & 0 & 48.70 & 19.13 & 37.6k \\
\texttt{llmlingua2-150w} & 146 & 47.65 & 16.52 & 40.3k \\
\bottomrule
\end{tabular}
\caption{All policy-handling configurations on the \textbf{retail} domain, ordered by pass@1. \textit{Pol./call} is the mean number of policy words sent per model call, plus the words the retriever or compressor itself reads where that applies. \textit{Words} is the mean total words per trajectory. $^{\dagger}$the \texttt{-Nw} suffix is the word budget we requested. These rows exceeded it, and \textit{Pol./call} is the length they actually sent.}
\label{tab:rag_compression_retail}
\end{table*}

\section{Turn Reward} \label{app:correctness_reward}
\subsection{Assistant Turn Reward}
For the assistant response turn, we have used the reward model as the LLM-as-a-judge. We prompt the LLM to score the generated response based on the ground truth response and penalize if the same information is not provided as the ground truth.

\begin{equation*}
    \mathcal{R}_{\text{assistant\_correct}} = 6 \cdot \frac{r}{10} - 3 \in [-3, 3]
\end{equation*}

where $r$ is the raw score from the reward model. The prompt we give the reward model is provided in \cref{app:prompt_turn_reward}.

\subsection{Tool-Call Turn Reward}
We adopt the reward formulation for tool-call turns from \citet{qian_toolrl_2025}. Unlike their approach, our method calls only a single tool at each turn, and we modify the reward score accordingly. The tool-call correctness reward consists of three components.

\vspace{2mm}
\textbf{a. Tool Name Matching:}

\begin{equation*}
    r_{\text{name}} = \frac{|G \cap P|}{|G \cup P|} \in [0, 1]
\end{equation*}
where \( G \) and \( P \) are the sets of tool names extracted from the ground-truth and predicted tool calls, respectively. 

\vspace{2mm}
\textbf{b. Parameter Name Matching:}

\begin{equation*}
    r_{\text{param}} = \frac{|\text{param}(G) \cap \text{param}(P)|}{|\text{param}(G) \cup \text{param}(P)|} \in [0, 1]
\end{equation*}

where \( \text{param}(P_G) \) and \( \text{param}(P_P) \) represent the parameter names of the predicted and ground-truth tool calls, respectively.

\vspace{2mm}
\textbf{c. Parameter Value Matching:}

\begin{equation*}
\begin{split}
    r_{\text{value}} &= \sum_{k \in \text{param}(G)} \mathds{1}[G[k] = P[k]] \\
    &\quad \in [0, |\text{param}(G_j)|]
\end{split}
\end{equation*}

where \( G[k]] \) and \( P[k] \) represent the values of the parameters for the predicted and ground truth tool calls.
The total reward score is computed by finding the optimal matching between \( P \) and \( G \) to maximize the total match score:

Total reward for each tool-call is:
\begin{equation*}
    r_{\text{match}} = r_{\text{name}} + r_{\text{param}} + r_{\text{value}} \in [0, S_{\max}]
\end{equation*}
where \( S_{\max} = 1 + 1 + |\text{param}(G)|\) denotes the maximum possible score.

Finally, we normalize the tool-call turn reward to lie within the range $[-3, 3]$.
\begin{equation*}
    \mathcal{R}_{\text{tool\_correct}} = 6 \cdot \frac{r}{S_{\max}} - 3 \in [-3, 3]
\end{equation*}
where \( r \) denotes the current match score from the current generation. The final correctness reward \( \mathcal{R}_{\text{tool\_correct}} \) is the normalized reward for the matching process. 

\subsection{Total Turn Reward}

\begin{equation*}
    r_{match\_turn} = \begin{cases} 
        \mathcal{R}_{assistant\_correct} \in [-3, 3] & \\
        \qquad \qquad \qquad \text{if \texttt{assistant}} \\
        \mathcal{R}_{tool\_correct} \in [-3, 3] & \\
        \qquad \qquad \qquad \text{if \texttt{tool-call}}
    \end{cases}
\end{equation*}

\begin{equation*}
\begin{split}
    \mathcal{R}_{correct} &= \mathds{1}[\text{turn}[G] = \text{turn}[P]] \\
    &\quad \cdot r_{match\_turn} \in [-3, 3]
\end{split}
\end{equation*}

where $\text{turn}[G]$ and $\text{turn}[P]$ denotes the generated turn (assistant or tool call) and predicted turn respectively.

\begin{table*}[!t]
\centering
\begin{adjustbox}{width=\textwidth,center}
    \begin{minipage}[b]{\columnwidth}
    \centering
    \resizebox{0.9\columnwidth}{!}{
        \begin{tabular}{@{}lr@{}}
        \toprule
        \textbf{Parameter} & \textbf{Value} \\
        \midrule
        Agent Template & Hermes \\
        Train Type & Full \\
        Learning Rate & $1 \times 10^{-5}$ \\
        LR Scheduler & Cosine with Min LR \\
        Minimum LR & $1 \times 10^{-6}$ \\
        Warmup Ratio & 0.05 \\
        Training Epochs & 15 \\
        Global Batch Size & 64 \\
        Max Sequence Length & 32,768 \\
        Max Gradient Norm & 1.0 \\
        Precision & bfloat16 \\
        \bottomrule
        \end{tabular}
    }
    \caption{SFT Configuration}
    \label{tab:sft_config}
    \end{minipage}
    
    \begin{minipage}[b]{\columnwidth}
    \centering
    \resizebox{0.9\columnwidth}{!}{
        \begin{tabular}{@{}lr@{}}
        \toprule
        \textbf{Parameter} & \textbf{Value} \\
        \midrule
        Agent Template & Hermes \\
        Train Type & Full \\
        Learning Rate & $1 \times 10^{-6}$ \\
        LR Scheduler & Cosine with Min LR \\
        Minimum LR & $1 \times 10^{-7}$ \\
        Warmup Ratio & 0.05 \\
        Training Epochs & 20 \\
        Global Batch Size & 512 \\
        Max Gradient Norm & 1.0 \\
        \midrule
        \multicolumn{2}{@{}l@{}}{\textit{GRPO-Specific}} \\
        \midrule
        KL Penalty Coefficient ($\beta$) & 0.1 \\
        Sampling Temperature & 1.0 \\
        Number of Generations & 4 \\
        Number of Iteration & 1 \\
        Max Completion Length & 2,048 \\
        \bottomrule
        \end{tabular}
    }
    \caption{GRPO Configuration}
    \label{tab:grpo_config}
    \end{minipage}
\end{adjustbox}
\end{table*}

\section{Implementation Details} \label{app:implementation_details}
For the \textbf{CoT generation pipeline}, we use VLLM \citep{kwon2023efficient} to run inference with \texttt{deepseek-r1} on 8 H200 GPUs, which takes approximately 8 hours. For evaluation, we employ \texttt{Claude-3.5-v1} via the Bedrock framework.

For \textbf{Multi-Stage Training}, we adopt the Swift framework \citep{zhao2025swift} to train all models. Training is conducted on 8 H200 GPUs with the following durations: 1 day for Stage~1, 8 hours for Stage~2, and 3 days for Stage~3. We initialize from a raw instruction-tuned model, fine-tune it for 15 epochs in Stages~1 and~2, and then apply GRPO training for 20 epochs.
In \cref{tab:sft_config,tab:grpo_config}, we provide the detailed hyperparameters used for our SFT and GRPO training. The same set of hyperparameters is applied across all model variants and sizes.

For \textbf{Evaluation}, we use 8 H200 GPUs with VLLM for agent-model inference and Bedrock for the user model. Unless otherwise specified, the user model is \texttt{Claude-4-Sonnet} with temperature~0.0, following the \taubench-framework.

For \textbf{Total Cost}, the durations above sum to roughly 896 H200-hours, 832 for the three training stages and 64 for CoT generation. The generation pipeline issues 5 LLM calls per CoT when it passes every rubric on the first attempt, and at most 29 when it uses all four refinement rounds: one generation and four rubric evaluations, then one error summary, one regeneration, and four evaluations in each round. Both costs are paid once. The recurring per-trajectory cost is reported in \cref{tab:compression_ratio} for our models and in \cref{tab:rag_compression} for the retrieval and compression baselines.

\section{Prompts} \label{app:prompts}
\subsection{Overriding Policy Prompt}
\begin{promptbox}[app:prompt_override_prompt]{Overriding Policy System Prompt}
\begin{Verbatim}[breaklines=true,breaksymbol=,fontfamily=tt]
### New policies:
Some policies have changed, below you are given the old policy with the new one. You MUST override the old policy with the new one.

<old_policy>
{{old_policy}}
</old_policy>

<new_policy>
{{new_policy}}
</new_policy>
\end{Verbatim}
\end{promptbox}

\subsection{CoT Trace Generation Prompts} \label{app:prompts_for_cot_data}

\begin{promptbox}[app:prompt_cot_gen]{CoT Generation}
\begin{Verbatim}[breaklines=true,breaksymbol=,fontfamily=tt]
You are an AI assistant generating a first-person Chain of Thought (CoT) explaining your reasoning for your response in a conversation. The reasoning MUST lead to the given response or function call.

Inputs:
- Business policy: {policy}
- Conversation history: {conversation}
- Your response: {assistant_response}

Task: Generate a first-person CoT explaining your reasoning for the last user query, connecting it to your final response or function call.

Steps:
1. Identify the last user query and your response
2. Analyze the business policy for relevant points needed for your response only (not future turns)
3. Formulate a first-person CoT explaining how policy informed your response

Required Properties:
- Completeness: Include all relevant policy parts for the response
- Atomicity: Be concise, focus only on last query and response
- Faithfulness: Only mention information explicitly stated in policy
- Style: First-person narrative mimicking natural thought

Analysis Process:
<reasoning>
1. State the exact last user query and your response
2. Quote relevant policy parts
3. List potential tools or function calls
4. Consider multiple policy interpretations
5. Identify edge cases or ambiguities
6. Connect each policy point to query and response
7. Consider potential misunderstandings in user query
8. Evaluate ethical implications
9. Draft initial first-person CoT
10. Check against required properties
11. Refine to directly connect request to final response
</reasoning>

Output Format:
<chain_of_thought>
[Write as natural thinking: "Okay, the user wants to... First I need to... Wait, the policy states... So I should... If [issue], then... Alright, here's what I'll do..."]
</chain_of_thought>

The CoT should be precise, focused only on the last turn, and adhere strictly to the policy information.
\end{Verbatim}
\end{promptbox}

\begin{promptbox}[app:cot_err_summary]{CoT Error Summary Generation}
\begin{Verbatim}[breaklines=true,breaksymbol=,fontfamily=tt]
You are an AI assistant analyzing and summarizing evaluation results for Chain of Thought (CoT) reasoning. Provide a concise summary of errors, reasons, and satisfied metrics to help improve CoT generation.

Inputs:
- Chain of Thought: {cot}
- Evaluation results: {eval_results}

Analysis Steps:
1. Read the CoT and all evaluation results
2. Analyze for errors, reasons, and satisfied metrics
3. Create detailed analysis in error_summary tags
4. Produce concise summary for AI model consumption

Output Format:
<error_summary>
Errors:
- [List errors with severity ratings 1-5, where 1=minor, 5=critical]
- [Quote relevant CoT parts for each error]

Reasons:
- [Explain primary reasons for errors]

Satisfied Metrics:
- [List satisfied evaluation metrics]

Improvement Suggestions:
- [Provide actionable suggestions]

Additional Notes:
[Any other relevant information]
</error_summary>

Requirements:
- Quote relevant CoT parts for each error
- Rate error severity (1-5 scale)
- Provide actionable improvement suggestions
- Structure for easy AI model comprehension
\end{Verbatim}
\end{promptbox}

\begin{promptbox}[app:prompt_cot_refine]{CoT Refinement Prompt}
\begin{Verbatim}[breaklines=true,breaksymbol=,fontfamily=tt]
You are an expert at refining Chains of Thought (CoT) for policy-based conversations. Analyze the inputs below and produce a refined CoT that fixes identified issues.

Inputs:
- Policy: {policy}
- Conversation: {conversation}
- Your last response: {assistant_response}
- Original CoT: {cot}
- Error summary: {error_summary}

Analysis Process:
<cot_refinement_process>
1. Identify relevant policy sections for the last user query
2. Extract relevant conversation context
3. List main issues from error summary
4. Plan how to address each issue
5. Consider policy implications and edge cases
6. Think through the refinement step-by-step in a natural, self-correcting manner
</cot_refinement_process>

Criteria:
- Completeness: Include all relevant policy parts
- Atomicity: Be concise, focus only on the last query/response
- Faithfulness: Only use information explicitly stated in policy
- Style: First-person narrative mimicking natural thought

Output Format:
<refined_cot>
[Write as if thinking aloud: "Okay, the user wants to... First I need to... Wait, the policy says... So I should... If [issue], then... Alright, here's what I'll do..."]
</refined_cot>

The refined CoT should read like an internal monologue addressing the last turn only.
\end{Verbatim}
\end{promptbox}

\begin{promptbox}[app:prompt_eval_completeness]{CoT Evaluation Rubric (Completeness)}
\begin{Verbatim}[breaklines=true,breaksymbol=,fontfamily=tt]
You are an expert AI evaluator assessing the completeness of a Chain of Thought (CoT) in relation to a policy and conversation. Evaluate how well the CoT captures all important policy information relevant to the assistant's immediate response to the last user query.

Inputs:
- Policy: {policy}
- Conversation: {conversation}
- Assistant's response: {assistant_response}
- Chain of Thought: {cot}

Task: Evaluate CoT completeness on a scale of 1-10. Completeness means including all important policy content necessary for the assistant's immediate response only (not future turns).

Evaluation Process:
1. Read policy, conversation, and CoT
2. Identify and quote relevant policy and conversation parts
3. List key policy points relevant to immediate response
4. Compare key points to CoT content
5. Note present or missing information
6. Consider impact on response quality
7. Provide specific action items for improvement

Rating Scale:
10: Perfect - All relevant information included
9: Near-perfect - Only minor details missing
8: Very good - Most important information included
7: Good - Important information included, some details missing
6: Above average - Key points covered, several details omitted
5: Average - Some important information, significant omissions
4: Below average - Major gaps in important information
3: Poor - Most important information missing
2: Very poor - Minimal relevant information
1: Inadequate - No relevant policy information

Output Format:
<cot_evaluation>
1. Policy review: [Summary of relevant policy points]
2. Conversation analysis: [Key aspects]
3. Relevant quotes: [Direct quotes relevant to immediate response]
4. Key policy points: [List of important points for immediate response]
5. CoT comparison: [Compare key points to CoT content]
6. Completeness assessment: [Identify included and missing information]
7. Impact analysis: [How completeness affects response quality]
8. Action items: [Specific improvements]
</cot_evaluation>

<analysis>
[Brief explanation of completeness level, strengths, and weaknesses]
</analysis>

<rating>
\boxed{[1-10]}
</rating>
\end{Verbatim}
\end{promptbox}

\begin{promptbox}[app:prompt_eval_faithfulness]{CoT Evaluation Rubric (Faithfulness)}
\begin{Verbatim}[breaklines=true,breaksymbol=,fontfamily=tt]
You are an expert evaluator assessing the faithfulness of a Chain of Thought (CoT) to a given policy. Faithfulness measures how closely the CoT adheres to information explicitly stated in the policy.

Inputs:
- Policy: {policy}
- Conversation: {conversation}
- Your response: {assistant_response}
- Chain of Thought: {cot}

Task: Evaluate CoT faithfulness on a scale of 1-10. Focus only on policies needed for the assistant's immediate response (not future turns or general context).

Definition:
A faithful CoT includes only information directly mentioned or immediately inferable from the policy.

Rating Scale:
1: Completely unfaithful - Mostly non-policy information
5: Moderately faithful - Mix of policy-based and non-policy information
10: Perfectly faithful - Only policy information

Evaluation Process:
1. Read policy, conversation, and CoT
2. Break down evaluation in cot_evaluation tags:
   a. List key points in CoT
   b. Quote relevant policy parts or note absence for each point
   c. Calculate percentage of faithful points
   d. Consider depth and accuracy of policy interpretation
3. Provide analysis
4. Give numeric rating
5. Include specific action items for improving faithfulness

Output Format:
<cot_evaluation>
[Detailed evaluation process]
</cot_evaluation>

<analysis>
[Concise analysis of CoT faithfulness to policy]
</analysis>

<rating>
\boxed{[1-10]}
</rating>
\end{Verbatim}
\end{promptbox}

\begin{promptbox}[app:prompt_eval_atomicity]{CoT Evaluation Rubric (Atomicity)}
\begin{Verbatim}[breaklines=true,breaksymbol=,fontfamily=tt]
You are an expert AI evaluator assessing the Atomicity of a Chain of Thought (CoT). Atomicity refers to selecting concise and important policy content directly relevant to the assistant's immediate response to the last user query.

Inputs:
- Policy: {policy}
- Conversation: {conversation}
- Your response: {assistant_response}
- Chain of Thought: {cot}

Task: Evaluate CoT Atomicity on a scale of 1-10. Focus only on policies needed for the assistant's immediate response (not future turns or general context).

Rating Scale:
1: Completely irrelevant - No relevant policy information
2: Highly irrelevant - Mostly irrelevant with tiny fraction relevant
3: Mostly irrelevant - Some relevant but overwhelmed by unnecessary details
4: Somewhat irrelevant - More irrelevant than relevant
5: Balanced but unfocused - Equal mix, lacking clear focus
6: Somewhat relevant - More relevant than irrelevant, some unnecessary details
7: Mostly relevant - Mostly relevant with few unnecessary details
8: Highly relevant - Almost all relevant, minimal extraneous content
9: Nearly perfect - Only relevant with perhaps one minor unnecessary detail
10: Perfect atomicity - Only most important and relevant information, perfectly concise

Evaluation Process:
<cot_evaluation>
1. Quote relevant policy parts for immediate response
2. Count relevant policy quotes
3. List key elements of last user query
4. Count key elements in user query
5. Match policy elements to query elements (only for immediate response)
6. Identify unnecessary or redundant information in CoT
7. Calculate percentage of relevant information
8. List missing key elements
9. Assess how well CoT leads to final response
10. Rate each aspect (1-10) with justification:
    a. Relevance of selected policy information
    b. Completeness of addressing query elements
    c. Absence of unnecessary information
    d. Contribution to final response
11. Provide specific action items for improving atomicity
</cot_evaluation>

<analysis>
[Concise analysis of CoT atomicity]
</analysis>

<rating>
\boxed{[1-10]}
</rating>
\end{Verbatim}
\end{promptbox}

\begin{promptbox}[app:prompt_eval_style]{CoT Evaluation Rubric (Style)}
\begin{Verbatim}[breaklines=true,breaksymbol=,fontfamily=tt]
You are an expert evaluator assessing the Style of an AI-generated Chain of Thought (CoT). Style measures whether the CoT uses first-person narrative that mimics natural thought processes with logical flow.

Inputs:
- Policy: {policy}
- Conversation: {conversation}
- Your response: {assistant_response}
- Chain of Thought: {cot}

Task: Evaluate CoT Style on a scale of 1-10.

Rating Scale:
1: Direct copy from policy, no narrative elements
2: Mostly extracted content, minimal original phrasing
3: Some narrative attempt, heavily reliant on policy wording
4: Basic narrative structure, lacks coherence and flow
5: Clear narrative attempt, occasional lapses into direct extraction
6: Consistent narrative style, could improve logical progression
7: Good narrative flow with clear reasoning, minor improvements needed
8: Strong narrative style with logical and coherent thought process
9: Excellent narrative with clear reasoning and smooth transitions
10: Exceptional narrative style, superior logical flow and original insights

Evaluation Process:
<cot_evaluation>
1. Quote CoT parts demonstrating narrative style or lack thereof
2. Analyze logical flow by numbering each thought step
3. Identify unnecessary elements or direct policy extractions
4. Discuss alignment with policy and conversation context
5. Compare CoT content with policy and conversation for accuracy and relevance
6. Provide specific action items for improving style
</cot_evaluation>

<analysis>
[Concise analysis of CoT style]
</analysis>

<rating>
\boxed{[1-10]}
</rating>
\end{Verbatim}
\end{promptbox}

\subsection{Reward Prompts} \label{app:prompts_reward}

\begin{promptbox}[app:prompt_user_req_to_policy]{User Request to Required Policies}
\begin{Verbatim}[breaklines=true,breaksymbol=,fontfamily=tt]
You are an AI tasked with identifying which policies are relevant to a given user request in a conversation. You will be provided with a list of atomic policies, a conversation history, and the last user request. Your job is to determine which policies need to be considered when formulating a response to the user's request.

First, here is the list of atomic policies:

<policies>
{policies}
</policies>

Now, here is the conversation history between the user and the agent:

<conversation_history>
{conversation_history}
</conversation_history>

The last user request is:

<user_request>
{user_request}
</user_request>

To complete this task, follow these steps:

1. Carefully read and understand each policy in the list.
2. Review the conversation history to understand the context of the interaction.
3. Analyze the last user request in relation to the policies.
4. Identify which policies are directly relevant to responding to the user's request.
5. Remember that the agent does not deviate from the policies at all, so be very selective in choosing relevant policies.

When you have determined which policies are relevant, output only the IDs of those policies. Your output should be a comma-separated list of policy IDs, without any additional explanation or justification.

For example, if policies AP001, RP001, and RP002 are relevant, your output should look like this:
<answer>
AP001,RP001,RP002
</answer>

If no policies are relevant, output:
<answer>
None
</answer>

Important: Your final output should consist of only the <answer> tag containing either the list of policy IDs or "None". Do not include any explanations, reasoning, or additional text in your final output.
\end{Verbatim}
\end{promptbox}

\begin{promptbox}[app:prompt_cot_to_policy]{Generated CoT to Mentioned Policies}
\begin{Verbatim}[breaklines=true,breaksymbol=,fontfamily=tt]
You will be given a list of atomic policies and an agent-generated response. Your task is to identify which policies are mentioned in the response and output their corresponding IDs.

First, here is the list of atomic policies:
<policies>
{policies}
</policies>

Now, here is the agent-generated response that you need to analyze:
<response>
{response}
</response>

To complete this task, follow these steps:

1. Carefully read through the list of atomic policies and familiarize yourself with their content.

2. Read the agent-generated response thoroughly.

3. For each policy in the list, determine if its content is mentioned or alluded to in the response. Pay attention to both explicit mentions and implicit references to the policy's content.

4. Keep track of the IDs of the policies that are mentioned in the response.

5. After analyzing the entire response, compile a list of the policy IDs that were mentioned.

Output your findings in the following format:
<mentioned_policies>
[List the IDs of the mentioned policies here, separated by commas]
</mentioned_policies>

If no policies were mentioned in the response, output:
<mentioned_policies>
None
</mentioned_policies>

Important: Your final output should consist of only the <mentioned_policies> tag containing either the list of policy IDs or "None". Do not include any explanations, reasoning, or additional text in your final output.
\end{Verbatim}
\end{promptbox}

\begin{promptbox}[app:prompt_cot_to_hal_policy]{Generated CoT to Hallucinated Policies}
\begin{Verbatim}[breaklines=true,breaksymbol=,fontfamily=tt]
You will be given a list of atomic policies and an agent-generated response. Your task is to identify any policies mentioned in the response that are not present in the given list of atomic policies. These are considered hallucinated policies.

First, here is the list of atomic policies:
<atomic_policies>
{policies}
</atomic_policies>

Now, here is the agent's response:
<agent_response>
{response}
</agent_response>

To complete this task, follow these steps:
1. Carefully read through the list of atomic policies.
2. Read the agent's response thoroughly.
3. Identify any policy statements or references in the agent's response.
4. Compare each identified policy in the response to the list of atomic policies.
5. If a policy mentioned in the response is not present in the atomic policies list, consider it a hallucinated policy.

For each hallucinated policy you identify, output it on a new line. The format should be as follows:
<hallucinated_policies>
hallucinated policy 1
hallucinated policy 2
</hallucinated_policies>

If there are no hallucinated policies, output:
<hallucinated_policies>
None
</hallucinated_policies>

Important: Your final output should consist of only the <hallucinated_policies> tag containing either the list of hallucinated policies or "None". Do not include any explanations, reasoning, or additional text in your final output.
\end{Verbatim}
\end{promptbox}

\begin{promptbox}[app:prompt_turn_reward]{Turn Reward}
\begin{Verbatim}[breaklines=true,breaksymbol=,fontfamily=tt]
You are given one ground truth response and one predicted response. You have to evaluate the predicted response based on completeness, correctness. The predicted response does not have to be exactly the same as the ground truth response, but it should contain the same information, without any misinformation. Score the predicted response from 0 to 10.

Ground truth response:
{gt_response}

Predicted response:
{pd_response}

Please reason step by step, and put your final answer within \boxed{}.
\end{Verbatim}
\end{promptbox}

\begin{promptbox}[app:prompt_format_reward]{Template for Format Reward}
\textbf{Assistant turn}
\vspace{-2mm}
\begin{Verbatim}[breaklines=true,breaksymbol=,fontfamily=tt]
<think>Policy-recall based Thinking</think>
{{final-response}}
\end{Verbatim}
\textbf{Tool-Call turn}
\vspace{-2mm}
\begin{Verbatim}[breaklines=true,breaksymbol=,fontfamily=tt]
<think>Policy-recall based Thinking</think>
{{tool-call-json}}
\end{Verbatim}
\end{promptbox}

\section{Use of AI Assistance}
The authors used Cursor\footnote{\url{https://cursor.com}}
 during development and ChatGPT\footnote{\url{https://chatgpt.com/}}
 for proofreading and refining the final manuscript. All content provided to these tools was originally created by the authors.

\end{document}